\documentclass{article}

\PassOptionsToPackage{numbers,compress}{natbib}
\usepackage[preprint]{neurips_2025}

\usepackage[utf8]{inputenc}
\usepackage[T1]{fontenc}
\usepackage{hyperref}
\usepackage{url}
\usepackage{booktabs}
\usepackage{amsfonts}
\usepackage{amsmath}
\usepackage{amssymb}
\usepackage{mathtools}
\usepackage{amsthm}
\usepackage{nicefrac}
\usepackage{microtype}
\usepackage{xcolor}
\usepackage{graphicx}
\usepackage{subcaption}
\usepackage{multirow}
\usepackage{array}
\usepackage{enumitem}
\usepackage{xspace}
\usepackage{pifont}
\usepackage[capitalize,noabbrev]{cleveref}
\usepackage{tikz}
\usepackage{wrapfig}
\usetikzlibrary{decorations.pathreplacing, arrows.meta, positioning, shapes.geometric}
\usepackage{pgfplots}
\pgfplotsset{compat=1.18}

\theoremstyle{plain}

\theoremstyle{definition}

\theoremstyle{remark}

\newcommand{\methodname}{\textbf{Incantation}\xspace}
\newcommand{\cmark}{\ding{51}}
\newcommand{\xmark}{\ding{55}}

\definecolor{darkgreen}{HTML}{4CAF50}
\definecolor{lightgreen}{HTML}{C8E6C9}
\definecolor{medblue}{HTML}{5B9BD5}
\definecolor{lightblue}{HTML}{BBDEFB}
\definecolor{warmorange}{HTML}{E07B39}
\definecolor{lightpeach}{HTML}{FCE4D6}

\title{Incantation: Natural Language as the Action Interface\\
for Multi-Entity Video World Models}

\author{%
  \makebox[\textwidth][s]{\textbf{Shangwen Zhu$^{1*}$ \hfill Qianyu Peng$^{8*}$ \hfill Zhao Pu$^{1*}$ \hfill Zhilei Shu$^{3}$ \hfill Xiangrui Ke$^{6}$}} \\
  \makebox[\textwidth][s]{\textbf{Zhaohu Xing$^{7}$ \hfill Zizhao Tong$^{4}$ \hfill Zeqing Wang$^{5}$ \hfill Xinyu Cui$^{9}$ \hfill Zian Zheng$^{6}$}} \\
  \makebox[\textwidth][s]{\textbf{Huangji Wang$^{1}$ \hfill Jian Zhao$^{9}$ \hfill Yeying Jin$^{5}$ \hfill Fan Cheng$^{1\dagger}$ \hfill Ruili Feng$^{2,6\dagger}$}} \\[0.8em]
  \makebox[\textwidth][c]{\normalfont $^{1}$SJTU \quad $^{2}$NVIDIA \quad $^{3}$Matrix Team \quad $^{4}$PKU \quad $^{5}$NUS} \\
  \makebox[\textwidth][c]{\normalfont $^{6}$UWaterloo \quad $^{7}$HKUST \quad $^{8}$HKU \quad $^{9}$ZGCA} \\[1em]
  \makebox[\textwidth][c]{%
  Github: \textcolor{magenta}{\url{https://github.com/zhushangwen/Incantation}}
  } \\[0.4em]
  \makebox[\textwidth][c]{%
  Project Page: \textcolor{magenta}{\url{https://matrixteam-ai.github.io/pages/Incanation}}
  } \\[0.4em]
  \makebox[\textwidth][c]{
  Dataset Preview: \textcolor{magenta}{\url{https://huggingface.co/datasets/zhush/incantation-elden-ring-scenes}}
    }
}

\begin{document}

\maketitle

\renewcommand{\thefootnote}{\fnsymbol{footnote}}
\footnotetext[1]{Equal contribution.}
% \footnotetext[3]{Corresponding author.}
\footnotetext[2]{Corresponding authors.}
\renewcommand{\thefootnote}{\arabic{footnote}}

\vspace{-2.5em}
\begin{center}
    \includegraphics[width=1.0\textwidth]{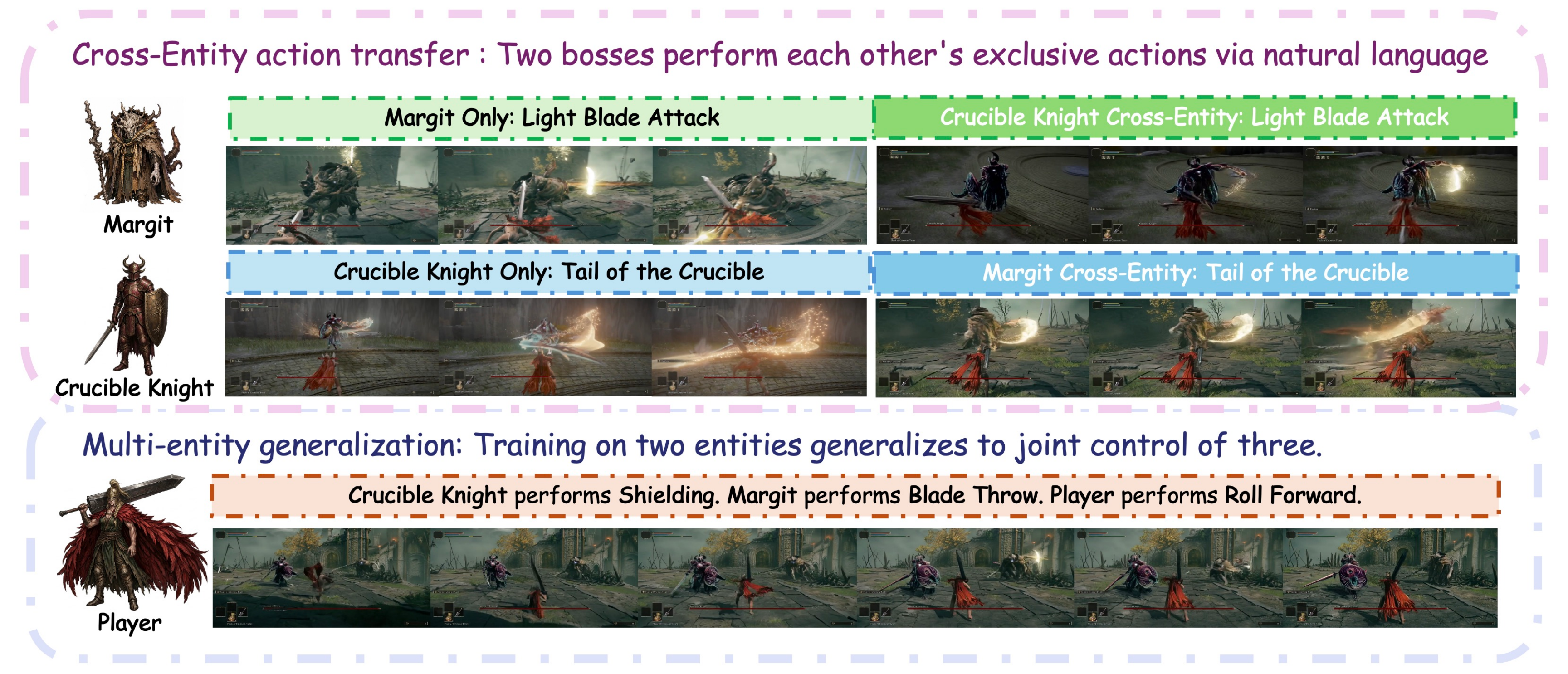}
    \vspace{0em}
    \captionof{figure}{\textbf{Demonstrations of \methodname's cross-entity action
    transfer and multi-entity control in the game \textit{Elden Ring}.} \textbf{\textit{(i)}} Two
    bosses, \textbf{Margit} and \textbf{Crucible Knight}, each possessing
    character-exclusive moves, are conditioned via natural language to perform each
    other's actions, each executed by both its native character and the other:
    \textbf{\textit{Light Blade Attack}} (Margit-exclusive,
    \textcolor{darkgreen}{green rows}) and \textbf{\textit{Tail of the Crucible}}
    (Crucible Knight-exclusive, \textcolor{medblue}{blue rows}), demonstrating
    \methodname's cross-entity generalization. \textbf{\textit{(ii)}} \methodname simultaneously
    controls three entities (two bosses and one player) each via a distinct
    natural-language prompt (\textcolor{warmorange}{orange rows}),
    trained on two-entity scenarios only.}
    \label{fig:demo}
\end{center}

\begin{abstract}
    Modern interactive video world models have achieved impressive 
    visual fidelity, yet lack fine-grained multi-entity control 
    and cross-entity, cross-world generalization. 
    We trace this gap to the \emph{action interface}:
    standard control protocols
    (e.g. animation IDs, device inputs, scene-level captions) 
    bind action semantics to specific entities or engines 
    at design time. 
    We propose \textit{natural language as the interface} to unlock 
    expressiveness that 
    no prior interface can achieve, and we present \methodname, 
    the first interactive video world model with per-latent-frame 
    ($0.25$\,s) natural-language conditioning that supports 
    simultaneous multi-entity control and \emph{concept-level} 
    cross-entity transfer beyond any fixed rendering pipeline. 
    We pair a pretrained bidirectional video backbone with 
    frame-local text cross-attention, and enable real-time 
    long-horizon streaming through ODE-initialized Self-Forcing 
    distillation with a RoPE-decoupled sliding KV-cache. 
    We surpass the Action-Index baseline on cross-entity transfer 
    ($89\%$ vs.\ $43\%$) and out-of-vocabulary prompts 
    ($90\%$ vs.\ $0\%$), and our $2$-step student sustains 
    $19.7$\,FPS at $480$p with stable FVD over 
    $2$-hour rollouts. 
    We further apply the same architecture and training recipe to
    \textit{The King of Fighters}, changing only the per-entity action vocabulary slots. 
    % We will release code, checkpoints, and a $128$-hour frame-accurate multi-entity action dataset, establishing 
    % natural language as a general, scalable action interface 
    % for video world models.
    We have released a preview subset of the \methodname dataset, containing manually collected Elden Ring player-boss combat clips with structured action-oriented metadata. Larger-scale Elden Ring and KOF data will be released with the full project.
\end{abstract}

\section{Introduction}
\label{sec:intro}

Modern video diffusion models~\citep{ho2022video, blattmann2023stable, wang2025wan} have driven a 
growing line of controllable interactive world models
~\citep{bruce2024genie, parkerholder2024genie2, parkerholder2025genie3, alonso2024diffusion, feng2024matrix, 
zhang2025matrixgame, he2025matrixgame2, valevski2024diffusion, oasis2024} to near-cinematic fidelity,
yet every such system inherits a structural limitation from the rendering pipelines it replaces:
actions are bound to engine-internal animation namespaces or device-level inputs, 
locking action semantics to a specific entity and engine.
This entity-and-engine binding forces a separate action vocabulary to be designed 
for every entity in every world,
making cross-entity and cross-world generalization an engineering burden rather 
than a modeling choice.
We argue that this is not an intrinsic property of multi-entity interactive video,
but a property of the \emph{action interface}: the protocol through which a user specifies what should
happen on the next frame, and replacing it fundamentally expands what such a model can express.
% Modern video diffusion models~\citep{ho2022video, blattmann2023stable, wang2025wan} now drive a growing line of controllable interactive world models~\citep{bruce2024genie, parkerholder2024genie2, parkerholder2025genie3, alonso2024diffusion, feng2024matrix, zhang2025matrixgame, he2025matrixgame2, valevski2024diffusion, oasis2024} to near-cinematic fidelity under a single shared camera, yet every such system inherits a structural limitation from the rendering pipelines it replaces: actions are bound either to engine-internal animation namespaces or to device-level inputs, both of which lock action semantics to a specific entity, body, and engine.
% Margit's animation~\#50, the Crucible Knight's animation~\#50, and animation~\#50 in any other engine refer to entirely unrelated events; the same is true of every keyboard mapping.
% As a consequence, an action defined for one entity cannot be issued to another, even when both share the scene, the camera, and the underlying motor primitives. We argue that this is not a property of multi-entity video as a problem, but a property of the \emph{action interface}: the protocol through which a user specifies what should happen on the next frame, and replacing it changes what such a model can express.

This bottleneck dominates in the \emph{single-viewpoint multi-entity} regime: 
a shared camera with two or more independently controllable entities, 
as in RPG (Role-Playing Game) combat and PvP (Player vs. Player) fighting.
This regime is central to competitive and adversarial gameplay, yet remains 
structurally underserved by interactive video world models.
Most controllable interactive world models confine control to a single entity,
leaving the rest as passive scenery~\citep{bruce2024genie, oasis2024, he2025matrixgame2, 
guo2025mineworld}, 
while recent 
multi-entity attempts sidestep the regime by dropping joint dynamics~\citep{po2026multigen}, 
abandoning the shared camera~\citep{solaris2025}, 
or controlling only one side~\citep{agarwal2026combat}.
None of these approaches admits a protocol with both fine-grained multi-entity control
and generalization across entities and worlds.
This shortfall ultimately traces back to the \emph{action interface} itself, 
which exhibits two conventional failure modes: 
\textbf{($\mathbf{1}$) Engine-internal animation labels} 
(per-world discrete IDs~\citep{alonso2024diffusion} and per-entity namespaces) 
bind each index to a specific animation at design time, 
so rendering any out-of-vocabulary (OOV) action is inherently inexpressible.
\textbf{($\mathbf{2}$) Human-device inputs}~\citep{bruce2024genie, parkerholder2024genie2,parkerholder2025genie3, oasis2024, he2025matrixgame2, guo2025mineworld,yu2025gamefactory, valevski2024diffusion} and 
\textbf{scene-level captions}~\citep{che2024gamegenx, tang2025hunyuan, team2026advancing} 
operate at the granularity of the player or the holistic scene rather than the individual entity,
thus lacking the critical per-entity addressability (e.g.,~non-player characters).
A viable multi-entity interface must therefore deliver both \emph{open-vocabulary semantics} 
for cross-entity semantic sharing and \emph{per-entity addressability} 
for independent, simultaneous control of each entity.

To address this limitation,
we propose \emph{a per-entity natural-language action interface} as the 
first to 
satisfy both desiderata, and present \methodname, 
the first interactive video world model 
supporting 
independent and simultaneous multi-entity control 
under a single shared viewpoint via 
per-frame natural-language conditioning
(\textbf{Throughout this paper, ``frame'' denotes a VAE-compressed latent frame 
unless otherwise specified; $\mathbf{1}$\,latent frame corresponds to $\mathbf{4}$\,pixel 
frames along the temporal axis; FPS denotes end-to-end pixel-frame throughput}).
Our interface assigns each entity its own 
syntactically isolated text segment within 
a shared prompt template at $0.25$\,s temporal 
granularity, enabling concurrent yet 
independent control of all entities.
Natural language shares semantics across entities by construction,
inherently allowing any action to be transferred from its 
native entity to another via a single textual phrase (Figure~\ref{fig:demo}).
We term this \emph{concept-level} cross-entity transfer: the model must 
synthesize both 
the motion and the visual concept on an entity that has no recording of the 
action, a capability inherently inaccessible to 
rendering pipelines bound to per-entity animation namespaces.
To our knowledge, no prior interactive video world model has explicitly addressed 
cross-entity action transfer at the level of per-frame, 
per-entity conditioning.

\begin{figure*}[t]
    \centering
    \captionsetup{justification=raggedright, singlelinecheck=false}
    \includegraphics[width=\linewidth]{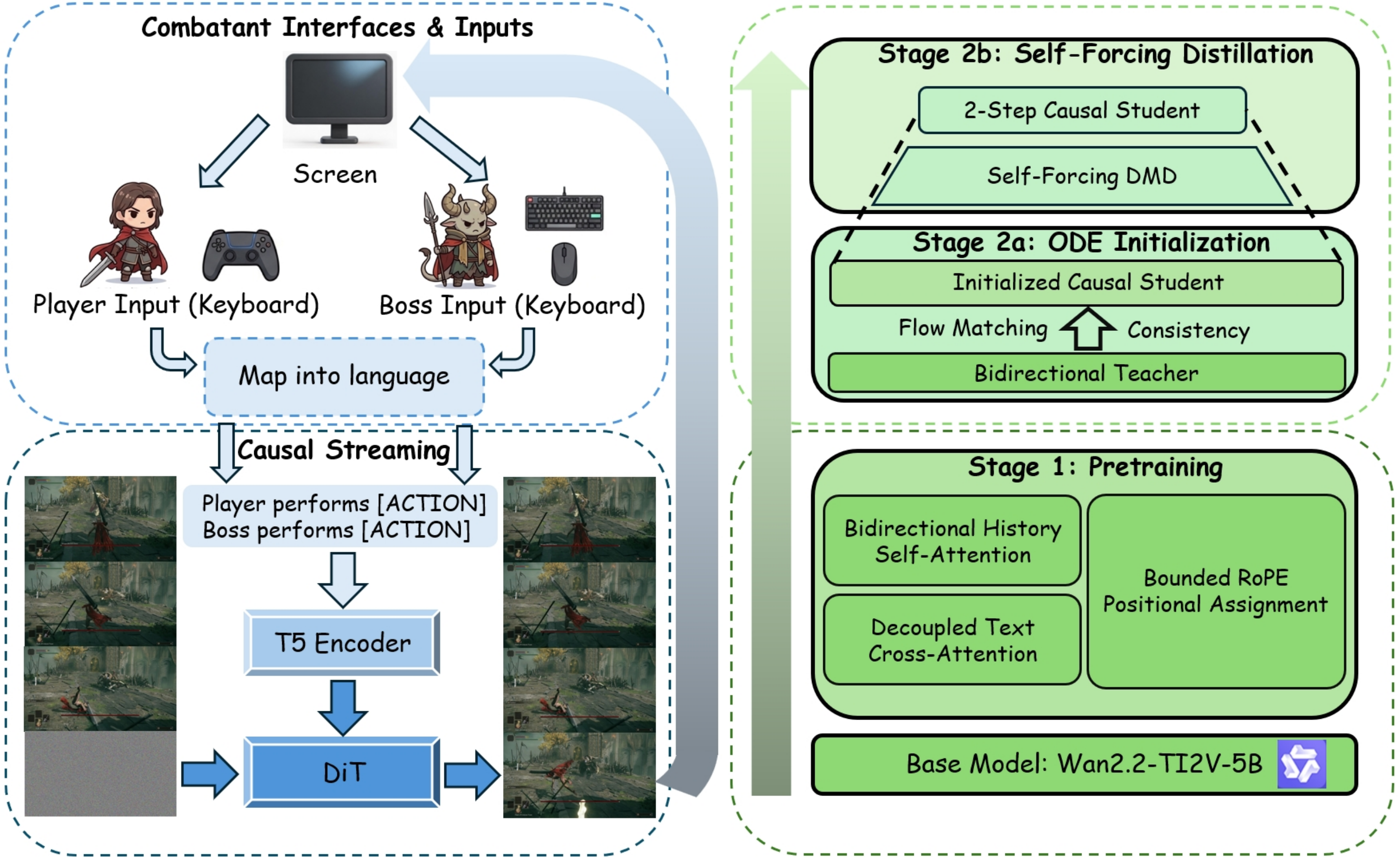}
    \caption{\textbf{Workflow of \methodname.}
    \textbf{Left:} \methodname translates combatant keyboard inputs into natural language prompts and autoregressively generates video frames in a causal streaming manner.
    \textbf{Right:} Training proceeds in two stages: ($1$)~Language-Conditioned Pretraining adapts the base model for per-frame text-driven generation; ($2$)~ODE-Initialized Self-Forcing Distillation enables real-time streaming via ODE-based flow matching initialization followed by Self-Forcing distillation.}
    \label{fig:workflow}
\end{figure*}

\methodname{} realizes this interface on top of a pretrained bidirectional video diffusion backbone~\citep{wang2025wan}. The core design is a \textbf{per-frame language-conditioned attention scheme}: decoupled text cross-attention is restricted exclusively to the noisy target frame and applied on top of bidirectional history self-attention, so each frame is steered by exactly its own action prompt without disturbing the backbone's pretrained priors or contaminating the committed history. We further enable real-time streaming inference by coupling ODE-initialized Self-Forcing distillation~\citep{chen2025selfforcing} with a RoPE-decoupled KV-cache sliding window, which collapses inference to two steps and keeps memory and positional geometry bounded over indefinite horizons.

Extensive experiments have demonstrated the structural advantage of \methodname{}'s 
natural-language interface. On cross-entity prompts (actions issued to entities 
that never executed them in training), \methodname{} attains $89\%$ Action 
Control Accuracy (ACA), far exceeding the $43\%$ of an Action-Index baseline 
whose accuracy merely tracks visual similarity rather than the action label 
itself. The gap widens to $90\%$ versus $0\%$ on OOV prompts, 
since the Action-Index interface cannot accept any prompt outside its fixed vocabulary. 
Besides its fine-grained per-frame control, \methodname{} sustains real-time long-horizon generation 
at $19.7$\,FPS with stable visual quality over $\mathbf{2}$-hour sessions, and replicates 
the performance on the visually unrelated \textit{King of Fighters} (KOF) world merely by 
vocabulary substitution alone, further validating its cross-world generalization capability. Our contribution can be summarized as follows:

\begin{enumerate}[leftmargin=*,topsep=1pt,itemsep=0pt]
\item We propose \textbf{natural language as the action interface for multi-entity video world models}, the first per-entity parallel control regime with open-vocabulary semantics, and demonstrate two structural capabilities unavailable to any discrete action-index, device-input, or scene-caption interface by construction: \textit{cross-entity action transfer} and \textit{out-of-vocabulary coverage}.
\item We present \methodname, the first interactive video world model with \textbf{per-frame, per-entity language conditioning} under a single shared viewpoint, achieving real-time multi-entity control for $\mathbf{>2}$ hours and reproducing its behavior on a second visually unrelated world under the same training recipe with vocabulary substitution as the only domain-specific change.
\item We construct a $128$-hour gaming dataset spanning two heterogeneous worlds (\textit{Elden Ring} and \textit{The King of Fighters}), the first dataset with \textbf{accurate per-frame, per-entity action labels} at $0.25$\,s temporal granularity, directly extracted from game memory at zero temporal offset.
\end{enumerate}

\section{Related Work}
\label{sec:related}

\paragraph{Interactive Video World Models.}
Most interactive video world models still simulate only a single controllable entity.
Following the world-model paradigm of~\citep{ha2018world, hafner2023mastering}, recent
diffusion-based engines such as GameNGen~\citep{valevski2024diffusion},
DIAMOND~\citep{alonso2024diffusion} and Oasis~\citep{oasis2024}, together with streaming
systems including the Genie series~\citep{bruce2024genie, parkerholder2024genie2, parkerholder2025genie3},
Matrix-Game~\citep{zhang2025matrixgame, he2025matrixgame2}, MineWorld~\citep{guo2025mineworld},
WorldPlay~\citep{sun2025worldplay}, Infinite-World~\citep{wu2026infiniteworld} and
Hunyuan-GameCraft-$2$~\citep{tang2025hunyuan}, all bind every action stream to one entity;
Vid2World~\citep{huang2025vid2world} and AVID~\citep{rigter2024avid} further repurpose
pretrained video diffusion models into action-conditioned world models under the same
single-agent setup.
Multi-entity attempts remain limited:
Solaris~\citep{solaris2025} synchronizes multi-player Minecraft videos but emits
per-player first-person streams rather than one holistic viewpoint, and
COMBAT~\citep{agarwal2026combat} renders a reactive Tekken~$3$ opponent inside a shared
view without any directable interface for its strategy;
ShareVerse~\citep{zhu2026shareverse} couples four agent-centric views on CARLA,
MultiGen~\citep{po2026multigen} enables editable multi-player rollouts via external
memory, and LiveWorld~\citep{duan2026liveworld} targets out-of-sight persistence, yet none delivers per-entity semantic commands.
Consequently, no existing system supports independent and simultaneous control
of multiple entities within one holistic scene.
\begin{figure*}[t]
    \centering
    \captionsetup{justification=raggedright, singlelinecheck=false}
    \includegraphics[width=\linewidth]{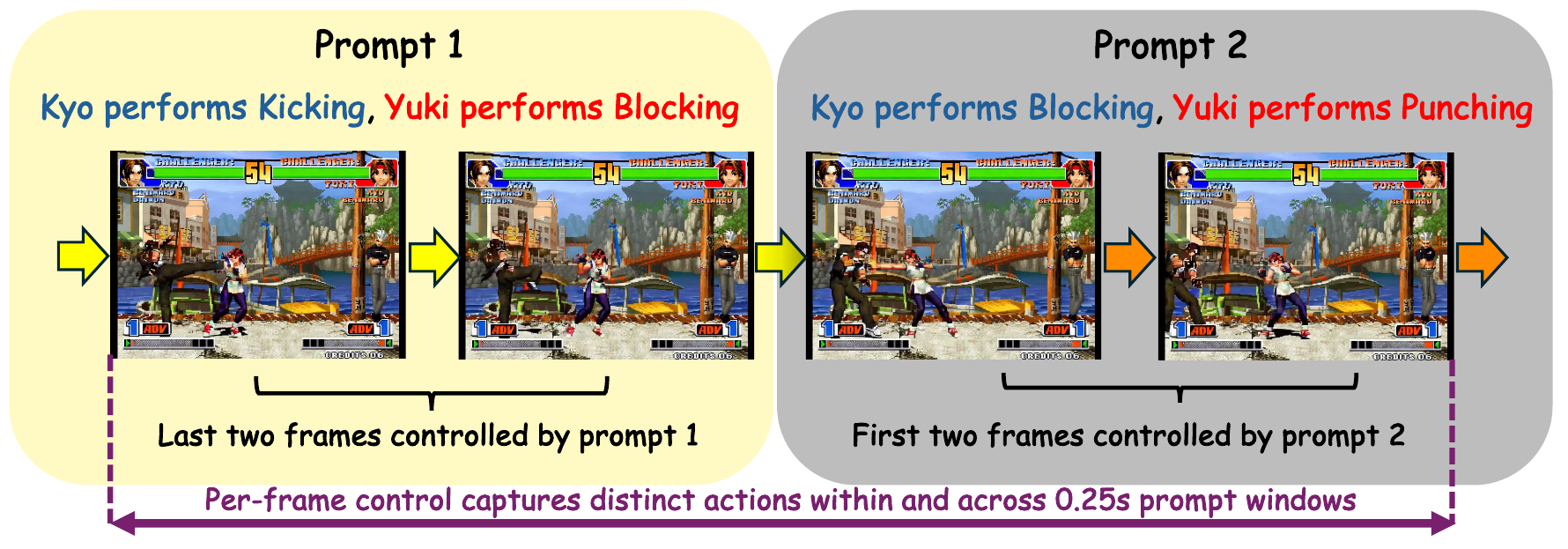}
    \caption{\textbf{Demonstrations of fine-grained multi-entity action control of \methodname in \textit{KOF}.}
    \methodname precisely responds to rapid action inputs and successfully captures
    actions as brief as $0.25$\,s (e.g., \textit{Punching}),
    demonstrating its fine-grained and responsive control capability.}
    \label{fig:kof_demo}
\end{figure*}
\paragraph{Action Interfaces of World Models.}
Existing world models inherit one of three action interfaces, each intrinsically limited
in generality and scalability across entities and worlds.
The first family encodes actions as \emph{engine-internal animation labels}, that is,
discrete identifiers exemplified by DIAMOND~\citep{alonso2024diffusion} on
Atari and Counter-Strike, where every index is bound at design time to a specific
in-game animation, leaving any out-of-vocabulary behavior inherently inexpressible.
The second family conditions generation on \emph{human-device inputs}, such as keyboard and mouse.
Representative systems include GameNGen~\citep{valevski2024diffusion},
the Genie series~\citep{bruce2024genie, parkerholder2024genie2, parkerholder2025genie3},
Oasis~\citep{oasis2024}, the Matrix-Game series~\citep{zhang2025matrixgame, he2025matrixgame2},
The Matrix~\citep{feng2024matrix}, MineWorld~\citep{guo2025mineworld},
WorldPlay~\citep{sun2025worldplay}, and GameFactory~\citep{yu2025gamefactory},
all of which condition on per-frame keyboard or mouse signals tied to a single player, so the schema cannot specify \emph{which} entity should act when
multiple entities co-exist within the scene.
The third family relies on \emph{scene-level captions}, where
GameGen-X~\citep{che2024gamegenx} feeds InstructNet with whole-clip multi-modal
instructions, Hunyuan-GameCraft-$2$~\citep{tang2025hunyuan} follows free-form prompts
such as ``open the door'', and LingBot-World~\citep{team2026advancing} further steers
global and local world events through textual prompts, each operating at the granularity
of the entire scene rather than any individual subject and thus conflating distinct
entities' behaviors under one global descriptor.
Across the three families, no prior interface simultaneously delivers open-vocabulary
semantics and per-entity addressability for independent simultaneous control of
multiple co-existing entities, exposing the core gap that our work targets.

\section{\methodname: Natural Language as the Action Interface}
\label{sec:method}

Realizing the language-as-action-interface end-to-end requires
addressing two architectural challenges inherent to any language-conditioned,
multi-entity interactive world model:
($1$) \emph{Per-frame language conditioning} and ($2$)
\emph{Real-time long-horizon streaming inference}.
We contribute one principled solution for each, structuring our pipeline into two stages.
\textbf{Stage~$\mathbf{1}$} (Section~\ref{sec:stage1}) addresses per-frame language conditioning
via a per-entity prompt formulation on a bidirectional backbone with decoupled
text cross-attention.
\textbf{Stage~$\mathbf{2}$} (Section~\ref{sec:stage2}) achieves real-time long-horizon
streaming generation through a two-stage distillation (ODE initialization followed by
Self-Forcing) combined with RoPE-decoupled KV-cache sliding.
Throughout this work, the action interface targets the
\emph{discrete-semantic action} regime, 
where each per-frame action
admits a textual description; 
% while
continuous control signals (e.g., camera
$\mathrm{SE}(3)$ trajectories) are out of scope and discussed in
Appendix~\ref{app:limitations}.

%% ═══════════════════════════════════════════════════════════════
\subsection{Stage 1: Language-Conditioned Architecture}
\label{sec:stage1}

We adopt natural language as the action interface, which inherently 
decouples the conditioning signal from any specific engine or entity and 
thereby enables \textbf{generalization} across both entity types and world domains. 
Realizing this interface on top of a pretrained bidirectional video 
backbone~\citep{wang2025wan} requires three coupled design choices: 
($1$) how multi-entity prompts are formulated, 
($2$) how attention is structured to turn 
high-level prompts into frame-accurate actions, and ($3$) how positional 
indices are assigned so that both training and bounded streaming inference 
stay in distribution.
% As discussed in Section~\ref{sec:intro}, device-level inputs are poorly suited to independently conditioning several heterogeneous entities, and discrete action IDs offer little semantic structure for cross-entity generalization. We therefore adopt natural language as the action interface. This subsection covers three coupled design choices: how prompts are formulated, how attention is structured to turn high-level prompts into frame-accurate actions, and how positional encodings are assigned so that both training and bounded inference stay in distribution.

\paragraph{Prompt Formulation. }
We represent multi-entity actions as a structured natural-language prompt with parallel, 
syntactically isolated slots (one per entity) at a $0.25$\,s granularity. As a concrete example, for two-entity control:
% Each 0.25\,s action window is described by a structured natural-language prompt with two parallel, syntactically isolated slots:
\begin{center}
  \small\texttt{Player performs [ACTION\_P]. Boss performs [ACTION\_B].}
\end{center}
This template supports both \textbf{simultaneous control} and 
\textbf{entity decoupling}: the temporal alignment of the two slots encourages the 
model to reason jointly about inter-entity dynamics within each frame, while the 
syntactic separation preserves independent control pathways for each entity. 
The template also extends naturally to settings with more or fewer entities 
by simply appending or omitting slots, requiring \textbf{no architectural modification}
and demonstrating the inherent \textbf{scalability} of the natural-language interface.

% \paragraph{Passive-Response Exclusion.}
% Passive physiological responses, in our setting taking-hit events, are not 
% deliberate actions issued by the controller, but involuntary responses 
% to external events happening to an entity. 
% Including them in the conditioning space introduces involuntary pauses that 
% corrupt the model's understanding of volitional movement. 
% We therefore exclude all such passive-response labels from the prompt vocabulary: the prompt vocabulary describes \emph{what an entity chooses to do}, not what happens to it.
% \paragraph{Excluding passive-response events from the prompt vocabulary.}
% Passive physiological responses---in our setting, taking-hit events---are not deliberate actions the controller issues but involuntary responses to events happening to an entity. Including them in the conditioning space introduces involuntary pauses that corrupt the model's understanding of volitional movement. We therefore exclude all such passive-response labels from the prompt vocabulary: the prompt vocabulary describes \emph{what an entity chooses to do}, not what happens to it.

\begin{figure*}[t]
    \centering
    \includegraphics[width=\linewidth]{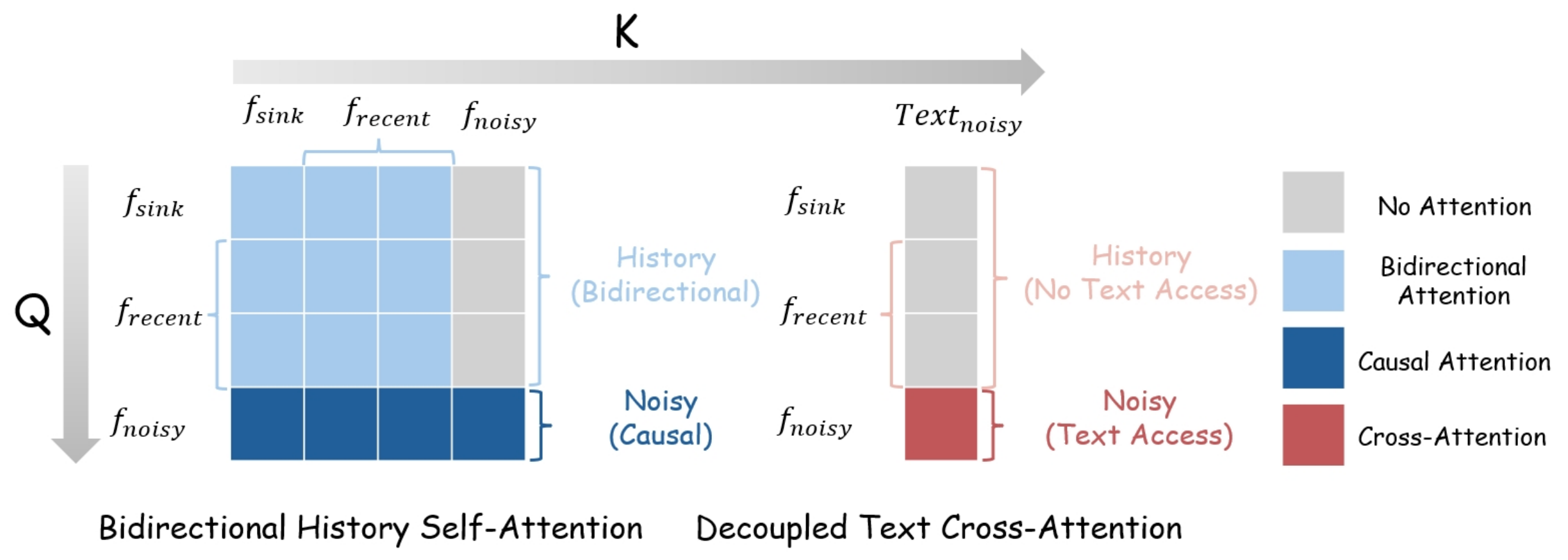}
    \caption{\textbf{Attention design.} Bidirectional self-attention is retained over history frames to preserve the spatio-temporal priors of the pretrained base model. Action cross-attention is restricted exclusively to the noisy target frame, preventing temporal cross-contamination. Together, these two constraints improve per-frame controllability without degrading generation quality.}
    \label{fig:masked_attention}
\end{figure*}

\paragraph{Context Assembly. }
In the autoregressive diffusion-based video generation framework,
each target frame is denoised by attending
to a context window of conditioning frames passed as clean latents.
We organize this window using a \emph{Sink + Recent + Noisy} context structure;
for each training step targeting frame~$t$:
% We adopt a \emph{Sink + Recent + Noisy} context structure. For each training step targeting frame~$t$:
\begin{itemize}[leftmargin=*,topsep=2pt,itemsep=1pt]
  \item \textbf{Sink frame} ($K_s{=}1$): the first frame of the episode, anchoring global context (arena geometry, character appearance) following the attention-sink mechanism of~\citet{xiao2023streamingllm}.
  \item \textbf{Recent frames} ($K_r{=}7$): the $7$ most recent clean latent tokens preceding $t$. Each latent token corresponds to $0.25$\,s of gameplay after the base model's VAE temporal compression, so the recent context spans $1.75$\,s of game time. We ablate $K_r$ in Appendix~\ref{app:stage2_ablation}.
  \item \textbf{Noisy target} ($K_n{=}1$): the partially-denoised latent of frame $t$.
\end{itemize}

\paragraph{Per-Frame Language-Conditioned Attention. }
The conventional approach, with causal self-attention over all visual tokens plus full text cross-attention, introduces two failure modes under per-frame language conditioning:
\textbf{($\mathbf{1}$)~Destruction of pretrained priors.} The Wan~$2.2$ base model was pretrained with full bidirectional attention; its weights encode symmetric co-occurrence statistics. Imposing a global causal mask discards these priors, requiring costly re-adaptation.
\textbf{($\mathbf{2}$)~Temporal cross-contamination.} Each action prompt $a_t$ describes exclusively what occurs at time~$t$. Allowing $a_t$ to cross-attend to history frames causes it to retroactively corrupt committed past representations, producing spurious action echoes in adjacent frames.
We address both issues with a dedicated attention mechanism for
per-frame language conditioning (Figure~\ref{fig:masked_attention}):
\textbf{($\mathbf{1}$) Bidirectional history attention.}
We apply \textit{full bidirectional self-attention} over the $(K_s + K_r)$ history tokens, preserving the base model's pretrained co-occurrence statistics. A causal boundary separates history from the noisy target, enforcing correct temporal ordering at generation time.
\textbf{($\mathbf{2}$) Decoupled text cross-attention.}
The per-frame action prompt $a_t$ cross-attends \emph{exclusively} with the noisy target token; history frames are masked out entirely. This prevents temporal cross-contamination: the current annotation cannot influence committed past representations. Ablation study appears in Appendix~\ref{app:stage1_ablation}.

\paragraph{Bounded RoPE Position Assignment.}
\label{sec:rope_design}
The naive sequential position assignment lets token indices grow
unboundedly during streaming inference, placing them outside the range
seen during training; this is a \textbf{RoPE out-of-distribution (OOD)} problem
that fundamentally breaks long-horizon generation.
We instead introduce two independent bounds: a sliding window size
$K_r$ (how many recent frames the KV cache holds) and a position cap
$C\ge K_r$ (the largest local RoPE index any token can receive).
The sink frame is permanently anchored at position $0$, the noisy
target at $\min(p_t, C)$ where $p_t$ is its absolute frame index, and
the $K_r$ recent frames occupy the consecutive positions immediately
preceding the target.
$K_r$ caps per-step compute and memory; $C$ caps the positional range
exposed to the model, and is set so every position used at inference
also occurs during training. Together the two prevent RoPE OOD and
enable the KV-cache sliding mechanism at inference (Section~\ref{sec:stage2}).
% The sink frame is assigned fixed position~0; recent frames receive sliding positions $1,\ldots,K_r$; the noisy target receives $K_r{+}1$. This scheme is \emph{bounded}: every token always carries a position in $\{0,\ldots,8\}$, the range seen during training. This guarantee serves double duty: it prevents RoPE OOD during training \emph{and} enables the KV-cache sliding mechanism at inference (Section~\ref{sec:stage2}).

\paragraph{Training Setup. }
We fine-tune Wan~$2.2$ TI2V-5B~\citep{wang2025wan} end-to-end on $16$ H100 GPUs
using Fully Sharded Data Parallel (FSDP) and mixed-precision training.
We employ a two-resolution curriculum: $1{,}000$ warmup steps at $256{\times}448$
(learning rate $2\!\times\!10^{-5}$), followed by $50{,}000$ steps at
$480{\times}832$ (learning rate $1\!\times\!10^{-5}$), with a global batch
size of $64$.
Training data are described in Section~\ref{sec:setup}.
% We fine-tune Wan~2.2 TI2V-5B~\citep{wan2025} end-to-end on 64$\times$H100 GPUs with FSDP and mixed precision. Training uses a two-resolution curriculum: 256$\times$448 for 1{,}000 warm-up steps (lr\,$=2\!\times\!10^{-5}$), then 480$\times$832 for 29{,}000 steps (lr\,$=1\!\times\!10^{-5}$), with global batch size 64. The training data is described in Section~\ref{sec:data_pipeline}.

%% ═══════════════════════════════════════════════════════════════
\subsection{Stage 2: Real-Time Streaming Inference}
\label{sec:stage2}

Real-time streaming inference is a prerequisite for any world model that aspires to support genuine interaction. The Stage~$1$ bidirectional teacher, however, requires $50$ denoising steps per frame and attends over a full visual context, neither of which is compatible with real-time play. Stage~$2$ addresses two coupled bottlenecks for this challenge: ($1$) reducing per-frame compute via distillation, and ($2$) bounding per-frame memory via KV-cache sliding while preserving positional coherence.
% The bidirectional teacher from Stage~1 requires ${\sim}$20 denoising steps per frame and attends over a full visual context, making real-time play infeasible. Stage~2 addresses two coupled bottlenecks: reducing per-frame compute via distillation, and bounding per-frame memory via KV-cache sliding while preserving positional coherence.

% \subsubsection*{ODE-Initialized Self-Forcing Distillation}
% We distill the teacher into a 2-step causal student in 
% two sub-stages: (1) ODE initialization to reconcile the bidirectional-to-causal
% attention mismatch, and (2) Self-Forcing for step reduction.
% We distill the teacher into a 2-step causal student in two sub-stages.

\paragraph{ODE Initialization Before Distillation.}
The teacher was pretrained with bidirectional history attention,
which grants rich spatio-temporal priors but is fundamentally incompatible with
the strictly causal attention required by streaming inference.
Before distillation, we must reconcile this mismatch. 
We initialize a causal student from the teacher's weights and align their 
predicted velocity fields via a flow-matching consistency
objective~\citep{lipman2022flow}:
\begin{equation}
  \mathcal{L}_{\text{ODE}} = \mathbb{E}_{\tau,v_0,\epsilon}\bigl[\|f_\theta(v_\tau;\,\text{causal}) - f_{\text{teacher}}(v_\tau;\,\text{bidir})\|_2^2\bigr].
\end{equation}
In practice,
this objective closes the attention-mask gap within $1{,}000$ steps at 
$480{\times}832$ resolution ($16$ H100 GPUs, learning rate\,$=5\!\times\!10^{-6}$, batch size $128$).
% \paragraph{ODE Initialization.}
% We initialize a causal student from the teacher's weights and align its predicted velocity field via a flow-matching consistency objective~\citep{lipman2022flow}:
% \begin{equation}
%   \mathcal{L}_{\text{ODE}} = \mathbb{E}_{\tau,v_0,\epsilon}\bigl[\|f_\theta(v_\tau;\,\text{causal}) - f_{\text{teacher}}(v_\tau;\,\text{bidir})\|_2^2\bigr]
% \end{equation}
% This resolves the attention-mask mismatch in 1{,}000 steps at 480$\times$832 (64$\times$H100, lr\,$=5\!\times\!10^{-6}$, batch size 128).

\paragraph{Self-Forcing Distillation.}
Building on the ODE-initialized student, 
we apply Self-Forcing~\citep{chen2025selfforcing} distillation to reduce inference to just
\textbf{$\mathbf{2}$ steps}. During training, the student conditions on \emph{its own
previously generated frames} rather than ground-truth frames, 
directly suppressing the compounding errors that 
would otherwise accumulate over autoregressive rollout.
% Starting from the ODE-initialized student, Self-Forcing~\citep{chen2025selfforcing} collapses inference to \textbf{2 steps}. During training the student conditions on its \emph{own} previously generated frames, directly suppressing the compounding errors that would otherwise accumulate during extended play.

\paragraph{RoPE-Decoupled KV-Cache Sliding Window. }
\label{sec:rope_decouple}
Under the bounded RoPE scheme in \Cref{sec:stage1} for OOD prevention, 
a bounded KV-cache sliding window is required to enable real-time streaming inference. 
However, the bounded relative positional indices are
time-dependent: 
after each eviction, surviving keys must be reassigned updated local 
relative positions. 
If RoPE-rotated keys are cached, their embeddings remain anchored to stale
indices and become inconsistent with the current query, causing temporal
flickering in the generated video. 
We therefore cache raw keys \emph{before} RoPE rotation and apply RoPE on-the-fly
with up-to-date local relative positions. 
Let $p^{\text{abs}}_i$ and $p^{\text{abs}}_t$ denote the absolute positions of 
cached frame $i$ and the current query $t$, respectively, with $C$ the local position cap defined in \Cref{sec:rope_design}.
Our local relative position assignment and RoPE-decoupled attention are:
\begin{align}
  p^{\text{local}}_i &= \text{clamp}\!\bigl(p^{\text{abs}}_i - \delta,\; 0,\; C\bigr),
    \quad \delta = \max\!\bigl(0,\, p^{\text{abs}}_t - C\bigr). \label{eq:local_pos} \\
  \text{Attn}(q_t, k_i) &= \text{Softmax}\{\bigl(q_t \cdot R(p^{\text{local}}_t)\bigr)
    \bigl(k_i^{\text{raw}} \cdot R(p^{\text{local}}_i)\bigr)^\top / \sqrt{d} \}.
  \label{eq:rope_decouple}
\end{align}
When the buffer is full, the oldest non-sink frame is evicted while the sink frame is
permanently retained at $p_{\text{sink}}^{\text{local}}{=}0$. The clamp cap $C$
keeps every local position within the range exercised during training,
ensuring long-horizon generation remains fully OOD-free. 
Together, our design guarantees: (a)~$\mathcal{O}(K_s + K_r)$ bounded memory;
(b)~all positions in-distribution; (c)~artifact-free evictions.
% Unbounded generation requires bounding VRAM while preserving positional coherence. Naive truncation causes flickering when evicted frames' absolute position indices suddenly disappear. We solve this by storing keys \emph{before} RoPE rotation and applying rotation on-the-fly using local positions:
% \begin{align}
%   p^{\text{local}}_i &= \text{clamp}\!\bigl(p^{\text{abs}}_i - \delta,\; 0,\; C\bigr), \quad \delta = \max\!\bigl(0,\, p^{\text{abs}}_{\text{current}} - C\bigr) \label{eq:local_pos} \\
%   \text{Attn}(q_t, k_i) &= \bigl(q_t \cdot R(p^{\text{local}}_t)\bigr)\bigl(k_i^{\text{raw}} \cdot R(p^{\text{local}}_i)\bigr)^\top / \sqrt{d}
%   \label{eq:rope_decouple}
% \end{align}
% When the buffer is full, the oldest non-sink frame is evicted; the sink entry is permanently retained. Because the sliding window bounds relative positions to $[0, K_r{+}1]{=}[0, 8]$---exactly the training range established by the RoPE design in Stage~1---long-horizon generation never encounters OOD positions. The local cap $C{=}16$ is a clamping bound; in practice, actual relative positions remain within $[0, 8]$. Together: (a)~VRAM is bounded to $\mathcal{O}(K_s + K_r)$; (b)~all positions stay in-distribution; (c)~evictions produce no visible artifacts.

\section{Experiments}
\label{sec:experiments}

Our experiments are organized around two research questions.
\emph{(i) With all else held equal, does the language interface 
offer capabilities unreachable by an Action-Index baseline?}
Section~\ref{sec:setup} introduces our testbed, baselines, 
and evaluation protocol; 
Section~\ref{sec:language_vs_id} then answers along three 
axes (in-distribution parity, cross-entity transfer, 
and out-of-vocabulary coverage), each designed to rule out a 
distinct confounder.
\emph{(ii) Does the same architecture sustain real-time 
inference and reproduce these gains in another 
visually unrelated world?}
Section~\ref{sec:system} addresses 
this by jointly reporting system-level metrics across 
\textit{Elden Ring} and \textit{The King of Fighters}.
\textbf{In addition, we conduct extensive ablation studies, with full results deferred to 
Appendix~\ref{app:stage1_ablation}--\ref{app:stage2_ablation} due to page constraints.}
% The experiments are organised around two questions. \emph{(i)~Does the language interface provide anything that a discrete-ID interface cannot, when nothing else differs?} Section~\ref{sec:setup} fixes the testbed, baselines, and metrics; Section~\ref{sec:language_vs_id} answers along three axes (in-distribution parity, cross-entity transfer, out-of-vocabulary coverage), each chosen to rule out a distinct confounder. \emph{(ii)~Does the same architecture run in real time and replicate the result on a second, visually unrelated world?} Section~\ref{sec:system} reports system-level metrics jointly across \textit{Elden Ring} and \textit{The King of Fighters}.

\subsection{Experimental Setup}
\label{sec:setup}

\paragraph{Testbed and Dataset.}
Our testbed spans two heterogeneous worlds: \textit{Elden Ring} 
($3$D action RPG, photorealistic) and \textit{The King of Fighters} 
(KOF; $2$D pixel-art). For \textit{Elden Ring}, we collect $30$\,h of Margit and $15$\,h 
of Crucible Knight boss-fight footage, 
with per-frame triplets $(v_t, a_t^{\text{player}}, a_t^{\text{boss}})$ read directly
from engine memory at zero temporal offset and 
player/boss vocabularies of $13$ and $47$ actions. 
For \textit{KOF}, we gather ${\sim}5{,}000$ $60$-second fighter-pair clips 
(${\approx}83$\,h)
% To enable cross-entity probing, we adopt a 
% \emph{mirrored naming} convention for the two bosses' signature tail attacks 
% (\texttt{Tail of the Margit} vs.\ \texttt{Tail of the Crucible}), 
% so the cross-entity prompt differs from its in-distribution counterpart by 
% exactly one word. 
(detailed in Appendix~\ref{app:setup_details}).
% \paragraph{Testbed and data.}
% The testbed spans two dissimilar worlds: \textit{Elden Ring} (3D action RPG, photorealistic) and \textit{The King of Fighters} (KOF; 2D pixel-art, side-scrolling). The Elden Ring training set comprises $30$\,h of Margit boss-fight gameplay and $15$\,h of Crucible Knight footage (used jointly for the cross-entity evaluation), with per-frame labels read directly from engine memory as zero-offset action triplets $(v_t, a_t^{\text{player}}, a_t^{\text{boss}})$; the joint player and boss vocabularies (Margit $\cup$ Crucible Knight, deduplicated) contain $13$ and $47$ actions respectively (per-entity decomposition in Appendix~\ref{app:vocab}). The KOF training set contains ${\sim}5{,}000$ $60$-second fighter-pair clips (${\approx}83$\,h). A \emph{mirrored naming} convention is applied to the two bosses' signature tail attacks (Margit: \texttt{Tail of the margit}; Knight: \texttt{Tail of the crucible}), so the corresponding cross-entity prompt differs from its in-distribution counterpart by a single token. Full data, vocabulary, and engine-instrumentation details are in Appendix~\ref{app:setup_details}.

\begin{figure*}[t]
    \centering
    \captionsetup{justification=raggedright, singlelinecheck=false}
    \includegraphics[width=\linewidth]{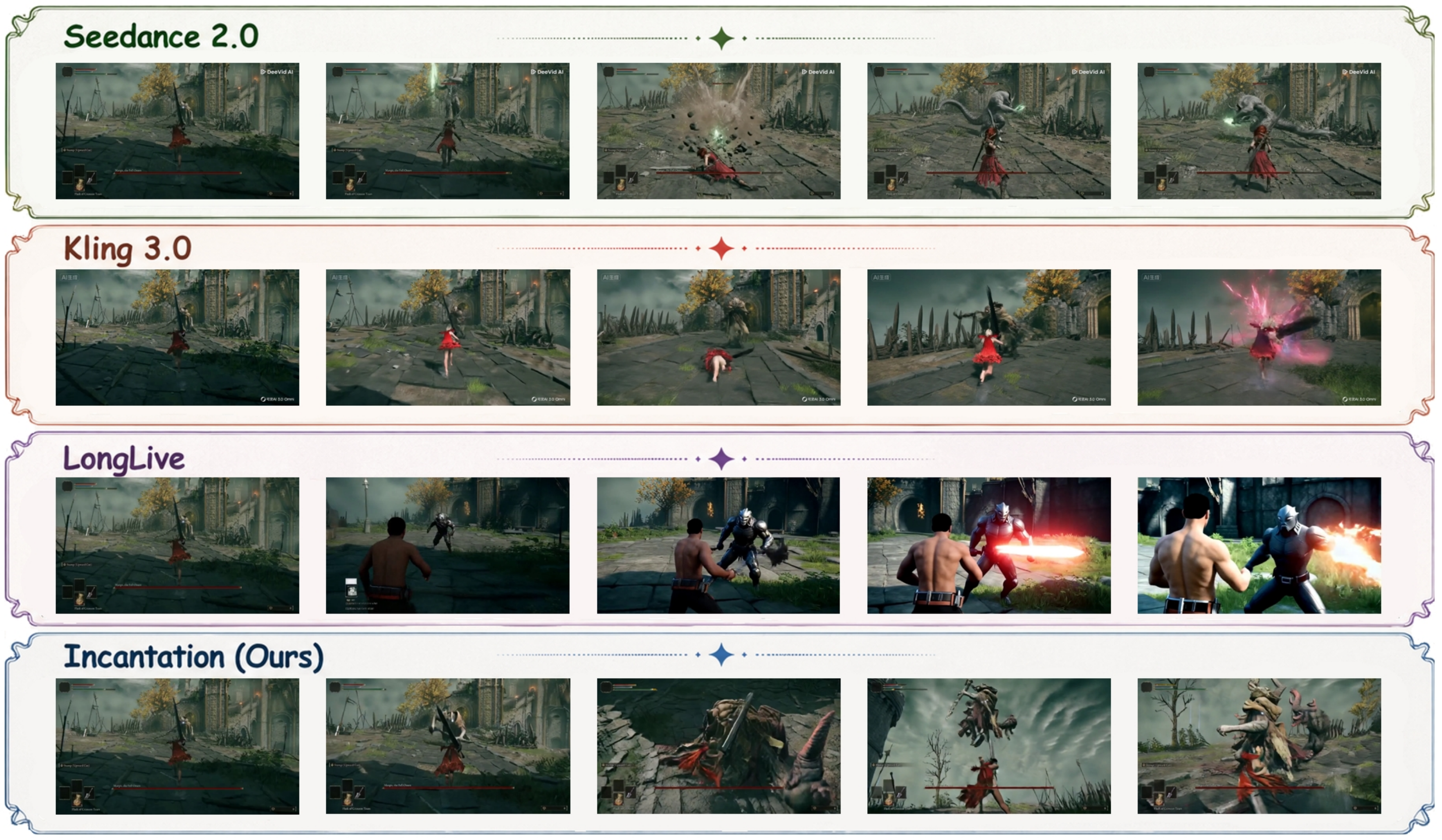}
    \caption{\textbf{Qualitative comparison of \methodname against leading video generation models on \textit{Elden Ring}.} \textbf{Seedance~$\mathbf{2.0}$}~\citep{seedance2026seedance20advancingvideo} and \textbf{Kling~$\mathbf{3.0}$}~\citep{kuaishou2026kling}
    achieve high visual fidelity yet fail on fine-grained player--boss interactions;
    \textbf{LongLive}~\citep{yang2025longlive} partially captures multi-entity dynamics but loses action fidelity and visual coherence.
    Only \methodname delivers precise per-frame multi-entity action control with genuine interactive modeling
    (prompts in Appendix~\ref{app:baseline_prompt}).
    \textbf{Existing world models are excluded as baselines, as none supports
    multi-entity modeling within a single holistic scene.}}
    \label{fig:baseline_comparison}
\end{figure*}
% \begin{wrapfigure}{r}{0.58\textwidth}
%   \centering
%   \captionsetup{justification=raggedright, singlelinecheck=false}
%   \includegraphics[width=\linewidth]{figures/fig_baseline_comparison.pdf}
%   \caption{\textbf{Qualitative comparison of \methodname against leading video generation models on \textit{Elden Ring}.} \textbf{Seedance~2.0}~\citep{seedance2026seedance20advancingvideo} and \textbf{Kling~3.0}~\citep{kuaishou2026kling}
%   achieve high visual fidelity yet fail on fine-grained player--boss interactions;
%   \textbf{LongLive}~\citep{yang2025longlive} partially captures multi-entity dynamics but loses action fidelity and visual coherence.
%   Only \methodname delivers precise per-frame multi-entity action control with genuine interactive modeling
%   (prompts in Appendix~\ref{app:baseline_prompt}).
%   \textbf{We exclude existing world models as baselines: their action interfaces are fundamentally incompatible
%   with our per-frame, per-entity natural-language conditioning paradigm.}}
%   \label{fig:baseline_comparison}
% \end{wrapfigure}

\paragraph{Baselines.}
We compare two conditioning variants that differ only in their conditioning pathways, with all other factors held \textit{identical}. The \textbf{Natural Language (NL) variant (ours)} encodes per-frame structured prompts via the model's pretrained text encoder into decoupled cross-attention layers, while the \textbf{Action-Index variant} instead represents each entity's action as a one-hot over the joint vocabulary, projected through a learnable linear layer. This capacity asymmetry is inherent, as equalizing it would artificially impose NL-level expressiveness into the Action-Index variant, rendering them fundamentally equivalent. Crucially, the joint vocabulary spans both entities, making cross-entity action indices technically injectable into either entity's context --- eliminating input-layer incompatibility as a confounding explanation for any cross-entity failure.  Thus, Action-Index should be viewed as a steel-man abstraction of common discrete control interfaces, including keyboard/controller inputs, animation IDs, and one-hot action tokens: it is stronger than raw low-level controls because it receives semantic action labels, and stronger than typical entity-local ID spaces because we use a shared joint vocabulary in which every transferred action remains addressable by a valid index. When it fails under cross-entity transfer, the failure therefore reflects the absence of compositional semantics in index-bound interfaces rather than lack of access to the target action. Both variants build on Wan~$2.2$ TI2V-5B~\citep{wang2025wan}, with causal-masked self-attention for real-time streaming inference. \textbf{As the first single-viewpoint multi-entity world model with per-frame, per-entity language actions, \methodname has no directly comparable baseline.}
% \paragraph{Baselines.}
% We compare two conditioning variants under \emph{identical} architecture, data, optimiser, batch size, and training compute. The \textbf{NL variant} (ours) feeds per-frame structured prompts through Wan~2.2's pretrained text encoder into the decoupled cross-attention of the noisy target. The \textbf{Discrete-ID variant} replaces only the conditioning pathway: each entity's action becomes a one-hot over the joint vocabulary defined in Section~\ref{sec:setup} ($|\mathcal{A}_{\text{boss}}^{\text{joint}}|{=}47$, $|\mathcal{A}_{\text{player}}^{\text{joint}}|{=}13$), projected through a learnable linear layer. The joint vocabulary makes cross-entity IDs technically injectable into either entity's context, so the ID variant cannot evade the cross-entity test through input-layer incompatibility. The base model is Wan~2.2 TI2V-5B~\citep{wang2025wan}; the causal student retains the same architecture with causal-masked self-attention.
\paragraph{Metrics and Protocols.}
Our primary metric is \textbf{ACA} (Action Control Accuracy), defined as the
fraction of generated clips judged consistent with their prompt by blinded annotators.
Owing to the absence of an established \textit{automated} evaluation 
protocol for this \textit{nascent} task, 
we adopt two rigorous blinded subjective evaluation protocols,
which better align with human perception,
to assess 
action control 
for compositional steering (Section~\ref{sec:language_vs_id}) and trajectory fidelity (Section~\ref{sec:system}), respectively.
Since these two protocols evaluate different aspects, their absolute
 ACA values are therefore not directly comparable.
Full evaluation details are provided in Appendix~\ref{app:setup_details}.
\subsection{Natural Language vs.\ Action-Index: Evidence Across Three Axes}
\label{sec:language_vs_id}

We evaluate whether natural language (NL) constitutes a genuinely superior
action interface over Action-Index through three controlled axes.
\textbf{Axis~$\mathbf{1}$} establishes a fair baseline by
confirming that NL and Action-Index perform comparably on actions seen during training,
ruling out model capacity or optimization as explanations for any subsequent gap.
\textbf{Axis~$\mathbf{2}$} tests whether NL generalizes
action semantics across different entities---ruling out memorization as the
source of any observed advantage.
\textbf{Axis~$\mathbf{3}$} exposes a structural limitation
of Action-Index by construction: NL can express any action through free composition,
whereas Action-Index cannot receive prompts outside its fixed vocabulary.
% \subsection{Language as the Action Interface}
% \label{sec:language_vs_id}

% We test the interface claim along three axes that each rule out a distinct potential confounder.
% \textbf{Axis~1} (in-distribution parity) rules out capacity or optimisation as the explanation for any subsequent gap.
% \textbf{Axis~2} (cross-entity semantic transfer) rules out memorisation.
% \textbf{Axis~3} (out-of-vocabulary coverage) is structural by construction: NL accepts any well-formed sentence, while ID has no input slot for any out-of-vocabulary prompt.

\paragraph{Axis $\mathbf{1}$: In-Distribution Parity. }
% \subsubsection{Axis 1: In-Distribution Parity}
% \label{sec:axis_indist}
We evaluate NL and Action-Index on the five most frequent actions in the training set,
which collectively dominate each entity's training data volume and ensure
strong supervision for both interfaces.
For each action, we report the ACA over $20$ trials under varied random setups. 
As presented in Table~\ref{tab:main_results},
NL leads Action-Index by $6$\,pp in aggregate on seen actions, thereby ruling out long-tail artifacts as a confounding explanation.
% We select the highest-frequency actions per entity by slot-0 count in the training manifest (actions for which both interfaces receive strong gradient signal) and report ACA($\geq$1) over $20$ trials per action under the prompt-injection protocol. The five actions retained collectively account for $74\%$ of the per-entity training-clip mass within the top-3 frequency buckets, ensuring the comparison is conducted on the densest part of each entity's training distribution rather than on long-tail moves.

\begin{table}[t]
  \caption{\textbf{Quantitative results of NL vs.\ Action-Index ACA across Axis~$\mathbf{1}$ and $\mathbf{2}$.}
    We report mean ACA over $20$ trials per action (Axis~$1$) and per action pair (Axis~$2$).
    NL outperforms Action-Index under both in-distribution and cross-entity settings,
    with a $6$\,pp advantage on seen actions and a $46$\,pp advantage on unseen cross-entity transfers.
    Full per-action and per-pair breakdowns are provided in
    Appendix~\ref{app:full_indist} (Table~\ref{tab:indist_full}) and
    Appendix~\ref{app:axis2_tiers} (Table~\ref{tab:cross_entity_full}), respectively.}
  \label{tab:main_results}
  \centering
  \setlength{\tabcolsep}{5pt}
  \begin{tabular}{p{6cm}cc}
    \toprule
    Evaluation Axis & \textbf{NL ACA (\%)} & Action-Index ACA (\%) \\
    \midrule
    Axis~$1$: In-Distribution Parity      & $\mathbf{95}$ & $89$ \\
    Axis~$2$: Cross-Entity Semantic Transfer & $\mathbf{89}$ & $43$ \\
    \bottomrule
  \end{tabular}
\end{table}

% \begin{table}[t]
%   \caption{\textbf{Axis~1: in-distribution controllability.} Highest-frequency actions per entity by slot-0 count in the training manifest (20 trials per action; three blinded annotators, median rating). \emph{Even on the densest part of each entity's training distribution, where memorisation effects most favour discrete IDs}, NL leads ID by $6$ percentage points on aggregate ($95\%$ vs.\ $89\%$), is non-inferior on every individual action, and strictly dominates on three of five. Per-trial $2/1/0$ score distributions are in Appendix~\ref{app:full_indist} (Table~\ref{tab:indist_full}).}
%   \label{tab:indist}
%   \centering
%   \setlength{\tabcolsep}{6pt}
%   \begin{tabular}{lcc}
%     \toprule
%     Action & \textbf{NL ACA (\%)} & Discrete-ID ACA (\%) \\
%     \midrule
%     Margit \textperiodcentered{} Double light blade throw & \textbf{95} & 85 \\
%     Margit \textperiodcentered{} Staff slam               & \textbf{80} & \textbf{80} \\
%     Knight \textperiodcentered{} Shield block             & \textbf{100} & \textbf{100} \\
%     Knight \textperiodcentered{} Overhead slash           & \textbf{100} & 85 \\
%     Knight \textperiodcentered{} Tail of the crucible     & \textbf{100} & 95 \\
%     \midrule
%     \textbf{Mean (5 actions)} & \textbf{95} & 89 \\
%     \bottomrule
%   \end{tabular}
% \end{table}

% NL leads ID by $6$\,pp on aggregate, strictly dominates on three of five actions, and is never out-performed; the lead is therefore established exactly where memorisation favours discrete IDs most, ruling out a long-tail-artefact explanation.

\paragraph{Axis $\mathbf{2}$: Cross-Entity Semantic Transfer. }
% \subsubsection{Axis 2: Cross-Entity Semantic Transfer}
% \label{sec:axis_cross}
To assess whether NL contributes semantic 
compositionality beyond the Action-Index interface, 
we examine whether the model can correctly 
interpret prompts for entity-action pairs that were
\emph{never} encountered during training. 
We test this on a hybrid model jointly trained on 
Margit and the Crucible Knight (disjoint action sets), evaluating 
five cross-entity action pairs. 
% Under the 
% mirrored naming convention defined in 
% Section~\ref{sec:setup}, 
Each cross-entity 
prompt differs from its in-distribution counterpart 
by a single entity-identity word (NL) or a 
one-hot index swap (Action-Index), ensuring that 
any performance drop reflects failed semantic 
generalization rather than exposure to 
unfamiliar vocabulary. We conduct $20$ trials 
per action pair and report mean ACA across all pairs. 
As shown in Table~\ref{tab:main_results}, 
NL outperforms Action-Index by $46$\,pp in mean 
ACA ($89\%$ vs.\ $43\%$), demonstrating that 
it is linguistic compositionality that 
enables robust cross-entity semantic transfer 
in a way discrete indexing fundamentally cannot. As an auxiliary automatic check, a VLM pairwise judge also favours NL on the same cross-entity pairs ($62\%$ vs.\ $37\%$ win rate), corroborating the human ACA trend (Appendix~\ref{app:axis2_tiers}, Table~\ref{tab:vlm_pairwise}).

\paragraph{Axis $\mathbf{3}$: Out-of-Vocabulary Coverage. }
% \subsubsection{Axis 3: Out-of-Vocabulary Coverage}
% \label{sec:axis_oov}

The third axis concerns prompts that extend, modify, or rephrase the training vocabulary 
while remaining compositionally meaningful
(e.g., \texttt{Double light blade throw} $\to$ \texttt{Dual light blade throw}). 
Here the NL-vs-Action-Index gap is \textit{structural} rather than quantitative: 
\emph{the Action-Index interface has no input slot for any such prompt}, so 
supporting any single one would require modifying the input-layer vocabulary 
fundamentally.
We construct four such probes, each with a single-word edit of one of the entity's 
top-$3$ frequent training prompts, giving Action-Index the strongest possible base 
embedding for a steel-man comparison (full set in Appendix~\ref{app:oov_probes}).  
Because no edit matches any predefined action index, the Action-Index interface scores exactly $\mathbf{0\%}$ 
regardless of model capacity, whereas NL achieves $\mathbf{90\%}$ aggregate ACA across 
the four probes in $40$ trials in total.
Stronger Action-Index baselines (e.g., factorized entity$\times$action tables) likewise reduce to either NL or our joint-vocabulary implementation (Appendix~\ref{app:stronger_id}), leaving the structural weakness of the Action-Index interface intact.
\textbf{Therefore, OOV coverage is unique to NL by construction: no scaling of an Action-Index interface can 
close this gap.}

% \subsubsection{Axis 3: Out-of-Vocabulary Prompts}
% \label{sec:axis_oov}

% The third axis concerns prompts that extend, modify, or rephrase the training vocabulary while remaining compositionally meaningful. Here the difference is structural rather than quantitative: \emph{the ID interface has no input slot for any such prompt}, and supporting any single one would require modifying the input-layer vocabulary itself. We evaluate four single-word edits of top-3 in-vocabulary base prompts (e.g.\ \texttt{Dual light blade throw.}, \texttt{Shield guard.}; full set in Appendix~\ref{app:oov_probes}). The corresponding base actions are among the highest-frequency for their entity, so the ID interface holds the strongest possible embedding for the original phrasing. Yet none of the probes corresponds to an index in the $47$-way joint vocabulary, so ID coverage is exactly $0\%$ regardless of the model's capacity. NL achieves $90\%$ aggregate ACA across the four probes ($40$ trials total). \textbf{OOV coverage is unique to NL by construction: no scaling of an ID-based interface can close it.} Stronger discrete-ID baselines (e.g.\ factorised entity$\times$action tables) collapse into either NL or our joint-vocabulary baseline (Appendix~\ref{app:stronger_id}); the gap therefore widens monotonically across the three axes ($+6$, $+46$, structural).

%% ═══════════════════════════════════════════════════════════════
\begin{table}[t]
  \caption{\textbf{Quantitative results across two visually unrelated worlds (Elden Ring, KOF).}
  With the same architecture and training recipe across worlds, the $2$-step student achieves $\mathbf{74/67\times}$ speedup over its teacher (Elden Ring/KOF), preserves ACA within $3$\,pp, and improves FVD.
  Seedance~$\mathbf{2.0}$ and LongLive are evaluated only by trajectory-conditioned ACA under the same $0.25$\,s per-entity labels; FVD/latency are omitted as non-comparable.
  Ablations and timing details appear in Appendices~\ref{app:stage1_ablation},~\ref{app:stage2_ablation}, and~\ref{app:realtime_pipeline}.}
  \label{tab:system}
  \centering\small
  \setlength{\tabcolsep}{6pt}
  \renewcommand{\arraystretch}{0.95}
  \begin{tabular}{llcccc}
    \toprule
    World & Model & Steps & FVD\,$\downarrow$ & ACA\,(\%)\,$\uparrow$ & Latency\,$\downarrow$ \\
    \midrule
    \multirow{4}{*}{Elden Ring}
      & Teacher (bidir.)               & $50$         & $206.2$          & $93.2$          & $12058.7 $\,ms/frame             \\
      & \textbf{Student (causal)}          & $\mathbf{2}$ & $\mathbf{138.6}$ & $\mathbf{90.4}$ & $\mathbf{163.4}$\,ms/frame \\
      & Seedance~$\mathbf{2.0}$~\citep{seedance2026seedance20advancingvideo} & --- & --- & $46.7$ & --- \\
      & LongLive~\citep{yang2025longlive} & 4 & --- & $20.3$ & $206.6$\,ms/frame \\
    \midrule
    \multirow{2}{*}{KOF}
      & Teacher (bidir.)               & $50$         & $170.1$          & $94.9$          & $10986.0$\,ms/frame             \\
      & \textbf{Student (causal)}          & $\mathbf{2}$ & $\mathbf{162.9}$ & $\mathbf{94.0}$ & $\mathbf{165.2}$\,ms/frame \\
    \bottomrule
  \end{tabular}
\end{table}
% \begin{wraptable}{r}{0.72\textwidth}   % r = 右侧，宽度自定义
%   \vspace{-1em}                         % 微调与上方文字的垂直间距
%   \caption{\textbf{System metrics across two visually unrelated worlds (Elden Ring, KOF).}
%   Across both worlds, the 2-step student achieves $\mathbf{50\times}$ speedup over its teacher
%   (160\,ms/frame on a single H100, real-time), preserves ACA within $3$\,pp, and \emph{improves} FVD--- all via pure vocabulary substitution with no architectural changes between worlds.
%   Stage~1/2 ablations in Appendices~\ref{app:stage1_ablation},\,\ref{app:stage2_ablation}.}
%   \label{tab:system}
%   \centering\small
%   \setlength{\tabcolsep}{4pt}           % 适当缩小列间距以适应窄宽度
%   \renewcommand{\arraystretch}{0.95}
%   \begin{tabular}{llcccc}
%     \toprule
%     World & Model & Steps & FVD\,$\downarrow$ & ACA\,(\%)\,$\uparrow$ & Latency\,$\downarrow$ \\
%     \midrule
%     \multirow{2}{*}{Elden Ring}
%       & Teacher (bidir.)           & 50         & 206.2          & 93.2          & 8\,s/frame             \\
%       & \textbf{Student (causal)}  & \textbf{2} & \textbf{138.6} & \textbf{90.4} & \textbf{160\,ms/frame} \\
%     \midrule
%     \multirow{2}{*}{KOF}
%       & Teacher (bidir.)           & 50         & 170.1          & 94.9          & 8\,s/frame             \\
%       & \textbf{Student (causal)}  & \textbf{2} & \textbf{162.9} & \textbf{94.0} & \textbf{160\,ms/frame} \\
%     \bottomrule
%   \end{tabular}
%   \vspace{-1em}
% \end{wraptable}

\subsection{Real-Time System and Cross-World Replication}
\label{sec:system}

% We retrain on KOF (2D pixel-art, side-scrolling, combo-based combat) under \emph{identical} architecture and hyperparameters as Elden Ring, changing only the action-vocabulary slots. Table~\ref{tab:system} compares the bidirectional teacher and the Self-Forcing causal $2$-step student at $480{\times}832$.
To validate the cross-world transfer of \methodname, we retrain on KOF under \emph{identical} architecture and 
hyperparameters as in Elden Ring, only modifying the action-vocabulary slots. 
Table~\ref{tab:system} compares the bidirectional teacher and its Self-Forcing causal $2$-step 
student at $480{\times}832$ on both worlds.
On Elden Ring, the student achieves a $\mathbf{74\times}$ speedup over the teacher at a comparable accuracy, 
while actually 
improving visual fidelity. 
On KOF, the same recipe under vocabulary substitution alone yields an 
analogous performance on a visually unrelated world. 
In addition, \methodname{} supports real-time streaming at $19.7$\,FPS end-to-end,
enabled by TAEHV~\citep{BoerBohan2025TAEHV}, a tiny VAE
(detailed in Appendix~\ref{app:realtime_pipeline}).
Although the training context spans only $1.75$\,s, 
the student maintains stable generation quality at much longer horizons: 
across continuous $30$- to $118$-minute sessions, 
FVD stays in a tight band (mean $166.0$, range $[162, 171]$) with no degradation 
over time (Appendix~\ref{app:long_horizon}). 
\section{Conclusion}
\label{sec:conclusion}
We present \methodname, the first interactive video world model to adopt natural language
as a \emph{per-frame, per-entity} action interface, overcoming the
expressiveness constraints of conventional interfaces. 
\methodname achieves
accurate multi-entity control in both cross-entity and out-of-vocabulary scenarios, and
sustains real-time streaming at $19.7$\,FPS over $2$-hour continuous horizons.
% \methodname further transfers seamlessly to a visually unrelated world
% by vocabulary substitution alone, showing the language interface
% generalizes across heterogeneous video worlds without architectural
% change.
\textbf{Limitations: }
% \textbf{($\mathbf{1}$) Data Scarcity.} Training requires accurate, fine-grained per-frame text
% annotations. While this is tractable in game scenarios via game memory access, extension
% \textbf{($\mathbf{1}$) Encoder Vocabulary Bound.} Open-vocabulary expressiveness is ultimately
% bounded by the pretrained text encoder. While composing concepts already embedded in the
% encoder generalizes naturally to unseen combinations, truly novel tokens absent from its
% training distribution require explicit encoder adaptation.
\textbf{($\mathbf{1}$) Annotation Channel.}
We read training labels from game memory because games offer
frame-accurate per-entity supervision at zero cost; this is a testbed
choice, not an interface property. The NL interface consumes per-entity
captions from any source---VLM auto-labelers, tele-operation logs, or
robot proprioception---without architectural change.
\textbf{($\mathbf{2}$) Continuous Controls.}
Our interface targets semantic actions; future hybrid controllers could combine language with continuous channels for precise camera $\mathrm{SE}(3)$ or force/velocity control. See Appendix~\ref{app:limitations} for details.

\begin{ack}
We thank the open-source community for tools enabling memory-accurate data collection. Elden Ring is a trademark of FromSoftware, Inc.\ and Bandai Namco Entertainment Inc.; this work is purely academic.
\end{ack}

\bibliographystyle{plainnat}
\bibliography{references}

\appendix
\newpage
\section{Appendix}
\label{sec:appendix}

{\small\setlength{\tabcolsep}{0pt}%
\begin{tabular}{@{}p{0.90\linewidth}r@{}}
  \hyperref[app:related_work]{\S\ref*{app:related_work}\enspace Comparison with Related Work}                                               & \pageref{app:related_work}    \\[2pt]
  \hyperref[app:limitations]{\S\ref*{app:limitations}\enspace Limitations and Future Work}                                                   & \pageref{app:limitations}     \\[2pt]
  \hyperref[app:qualitative_rollouts]{\S\ref*{app:qualitative_rollouts}\enspace Additional Qualitative Rollouts}                                                                    & \pageref{app:qualitative_rollouts}           \\[2pt]
  \hyperref[app:vocab]{\S\ref*{app:vocab}\enspace Full Action Vocabulary}                                                                    & \pageref{app:vocab}           \\[2pt]
  \hyperref[app:baseline_prompt]{\S\ref*{app:baseline_prompt}\enspace Baseline Prompt Settings}                                              & \pageref{app:baseline_prompt} \\[2pt]
  \hyperref[app:setup_details]{\S\ref*{app:setup_details}\enspace Experimental Setup Details}                                                & \pageref{app:setup_details}   \\[2pt]
  \hyperref[app:reliability]{\S\ref*{app:reliability}\enspace Annotator Reliability}                                                         & \pageref{app:reliability}     \\[2pt]
  \hyperref[app:stage1_ablation]{\S\ref*{app:stage1_ablation}\enspace Stage~$\mathbf{1}$: Conditioning Architecture Ablation}                & \pageref{app:stage1_ablation} \\[2pt]
  \hyperref[app:stage2_ablation]{\S\ref*{app:stage2_ablation}\enspace Stage~$\mathbf{2}$: KV-Cache and Bounded RoPE Ablation}                & \pageref{app:stage2_ablation} \\[2pt]
  \hyperref[app:stronger_id]{\S\ref*{app:stronger_id}\enspace Stronger Action-Index Baselines Collapse into NL}                              & \pageref{app:stronger_id}     \\[2pt]
  \hyperref[app:full_indist]{\S\ref*{app:full_indist}\enspace Axis~$\mathbf{1}$: Full In-Distribution Score Distributions}                   & \pageref{app:full_indist}     \\[2pt]
  \hyperref[app:axis2_tiers]{\S\ref*{app:axis2_tiers}\enspace Axis~$\mathbf{2}$: Full Cross-Entity Distributions and Per-Tier Mechanism}     & \pageref{app:axis2_tiers}     \\[2pt]
  \hyperref[app:oov_probes]{\S\ref*{app:oov_probes}\enspace OOV Probe Set}                                                                  & \pageref{app:oov_probes}      \\[2pt]
  \hyperref[app:failure]{\S\ref*{app:failure}\enspace Failure Case Analysis}                                                                 & \pageref{app:failure}         \\[2pt]
  \hyperref[app:long_horizon]{\S\ref*{app:long_horizon}\enspace Long-Horizon Stability}                                                      & \pageref{app:long_horizon}    \\[2pt]
    \hyperref[app:persistent_state]{\S\ref*{app:persistent_state}\enspace Extension: Persistent Entity State via an Observer--Tracker--Policy Loop} & \pageref{app:persistent_state}\\[2pt]
  \hyperref[app:realtime_pipeline]{\S\ref*{app:realtime_pipeline}\enspace Realtime Pipeline} & \pageref{app:realtime_pipeline}\\
\end{tabular}}\par\vspace{4pt}

\subsection{Comparison with Related Work}
\label{app:related_work}

\begin{table}[t]
  \caption{Systematic comparison of interactive video world models. 
  \cmark~=~supported, \xmark~=~not supported, $\sim$~=~partial. 
  \emph{Multi-entity} requires independent and simultaneous control of two 
  distinct entities. 
  \emph{Semantic NL} requires per-frame natural language action 
  conditioning.}
  \label{tab:comparison}
  \centering
  \begin{tabular}{lcccc}
    \toprule
    Model & Multi-entity & Semantic NL & $\geq$$16$FPS & $>$$5$min \\
    \midrule
    GameNGen~\citep{valevski2024diffusion}     & \xmark & \xmark & \cmark & \cmark \\
    Genie~$2$~\citep{parkerholder2024genie2}     & $\sim$ & \xmark & \cmark & \xmark \\
    Genie~$3$~\citep{parkerholder2025genie3}     & $\sim$ & \xmark & \cmark & $\sim$ \\
    The Matrix~\citep{feng2024matrix}          & \xmark & \xmark & \cmark & \cmark \\
    Matrix-Game $2.0$~\citep{he2025matrixgame2}  & \xmark & \xmark & \cmark & $\sim$ \\
    LingBot-World~\citep{team2026advancing}  & \xmark & \xmark & \cmark & \cmark \\
    Solaris~\citep{solaris2025}                & \cmark & \xmark & $\sim$ & $\sim$ \\
    \midrule
    \textbf{\methodname (Ours)}                & \cmark & \cmark & \cmark & \cmark \\
    \bottomrule
  \end{tabular}
\end{table}

Table~\ref{tab:comparison} presents a systematic comparison of 
\methodname against representative interactive video world models 
along four dimensions: multi-entity control, 
semantic natural language interface, 
real-time frame rate ($\geq$$16$FPS), and long-horizon 
generation ($>$$5$min).

\paragraph{Explanation of absence of world model baselines.}
As established in Table~\ref{tab:comparison} and 
Section~\ref{sec:related}, 
while a small number of existing world models 
accommodate concurrent multi-entity modeling to 
some extent, 
none achieves independent and simultaneous control 
of multiple entities within a single holistic scene. 
Because \methodname is, to our knowledge, 
the first system to address this setting, 
no directly comparable baseline exists for 
quantitative evaluation.

\paragraph{Additional Related Work for Efficient Streaming Video Generation. }
Several lines of work advance streaming video generation from complementary angles.
On the diffusion side, Flow Matching~\citep{lipman2022flow} and Distribution Matching Distillation (DMD)~\citep{yin2024one} substantially reduce the number of inference steps.
On the autoregressive side, Self-Forcing~\citep{chen2025selfforcing} eliminates exposure bias caused by the training-inference discrepancy.
For long-horizon memory management, StreamingLLM~\citep{xiao2023streamingllm} and LM-Infinite~\citep{han2023lminfinite} bound memory usage via attention-sink tokens and sliding-window KV caches.
Our work integrates these advances together to achieve real-time long-horizon streaming generation.

\subsection{Limitations and Future Work}
\label{app:limitations}

\methodname has three limitations that point toward concrete future directions.
First, our training labels are obtained by direct in-engine memory instrumentation, since games are the only domain that simultaneously offers frame-accurate, per-entity, zero-cost supervision at interactive frame rates; this is a deliberate testbed choice and is orthogonal to the language interface itself, which only consumes per-entity action captions and is agnostic to how those captions are produced. Closing the annotation channel for non-instrumented domains, namely real-world video and closed-source engines, reduces to producing per-entity captions from an alternative source such as vision--language auto-labelers, tele-operation logs, or robot proprioception, all of which integrate without architectural change. We therefore view this as a data-side problem rather than a structural restriction of \methodname.
Second, open-vocabulary expressiveness is ultimately bounded by the pretrained text encoder: composing concepts already within its training distribution generalizes naturally to unseen combinations, whereas truly novel tokens require explicit encoder adaptation. Our current interface also targets semantic action control rather than numerically precise continuous controls, such as camera $\mathrm{SE}(3)$ trajectories or force/velocity commands in robotic manipulation. This does not require replacing the language interface: a natural extension is to keep language as the high-level per-entity semantic channel and add a parallel continuous-control module, whose embeddings can be fused with the same per-frame conditioning layers used by \methodname.
Third, episode-level state beyond the generator's ${\sim}1.75$\,s context window is maintained by a hand-specified persistent-state module (Appendix~\ref{app:persistent_state}); replacing it with a learned, world-agnostic alternative remains open.
In strictly single-agent scenarios, the language interface also degenerates into a relabelling of discrete inputs and offers no representational advantage over conventional action identifiers.
Extending \methodname to non-game interactive-video domains, where per-entity annotations must be inferred rather than read from engine memory, constitutes the most immediate direction for future work.

\subsection{Additional Qualitative Rollouts}
\label{app:qualitative_rollouts}
\begin{figure}[h]
    \centering
    \includegraphics[width=\linewidth]{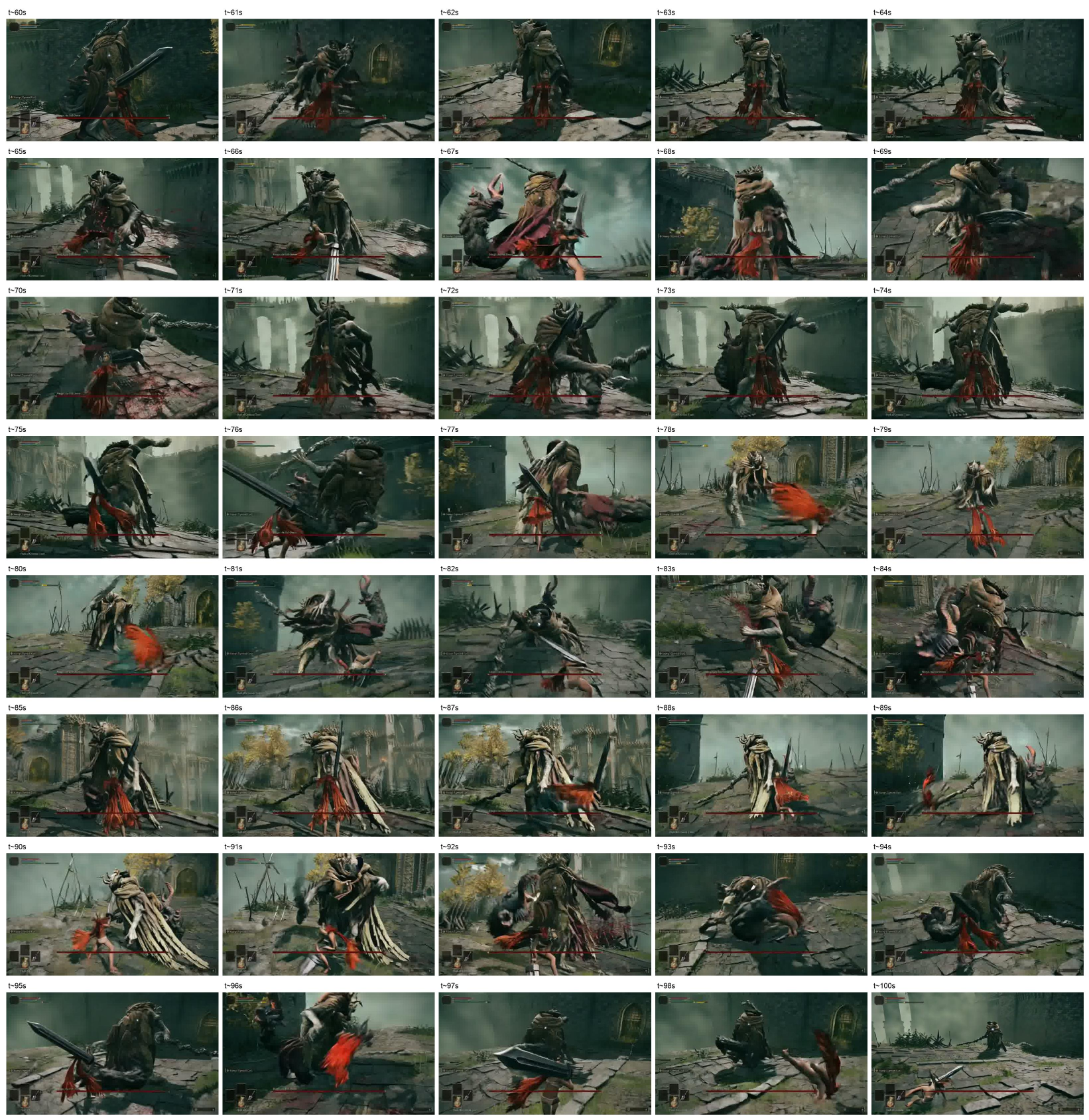}
    \caption{\textbf{Elden Ring rollout from a continuous Margit session.}
    We show $40$ frames sampled from the generated stream starting at the $1$-minute mark.
    The sequence illustrates long-horizon visual stability and fine-grained player--boss interaction in a complex 3D adversarial scene.}
    \label{fig:margit_rollout_1min}
\end{figure}

\begin{figure}[h]
    \centering
    \includegraphics[width=\linewidth]{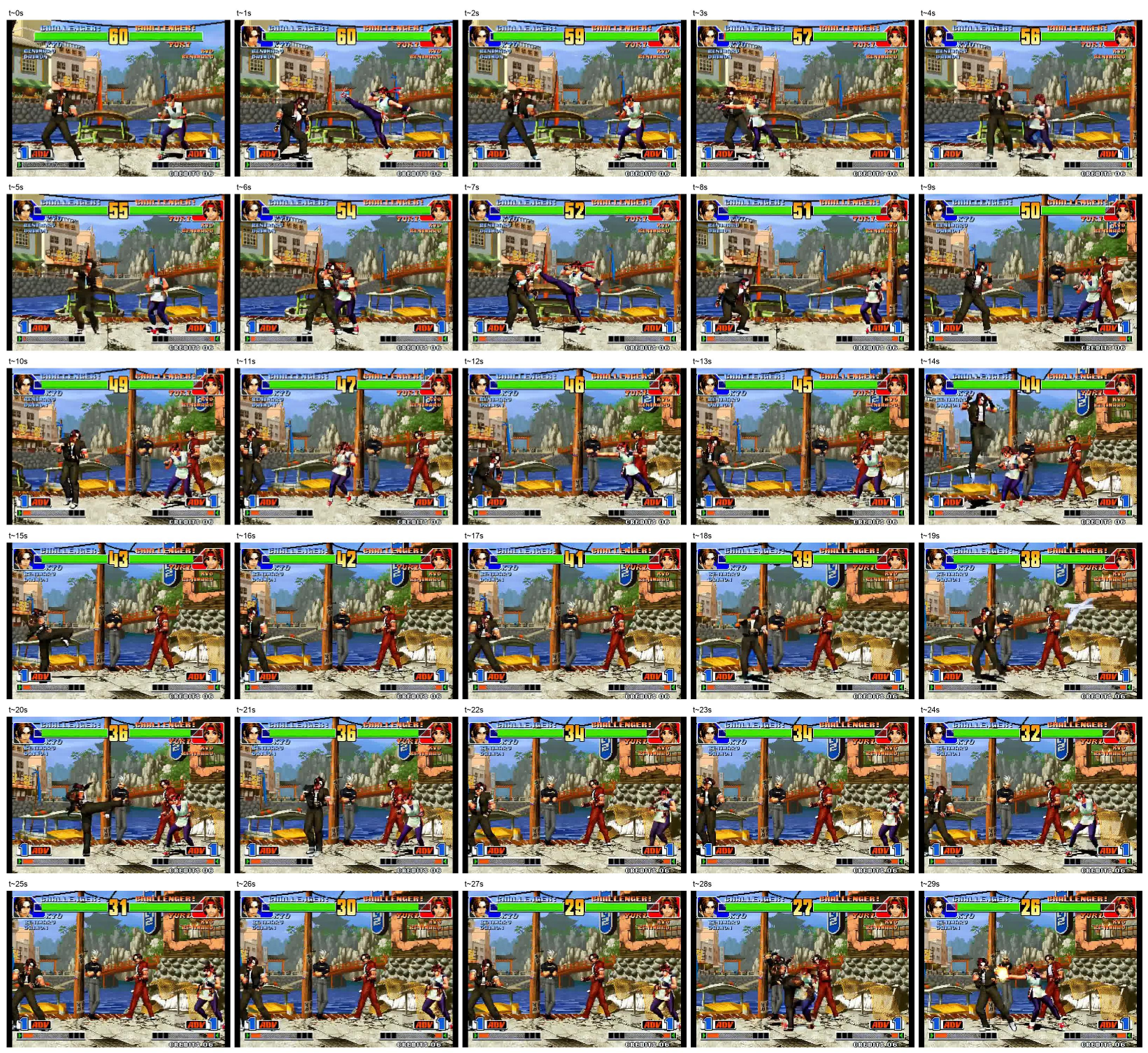}
    \caption{\textbf{KOF rollout under the same architecture and training recipe.}
    We show $30$ frames from a KOF rollout.
    The sequence illustrates that the same per-entity language-conditioning recipe also supports visually distinct 2D fighting gameplay.}
    \label{fig:kof_rollout_0}
\end{figure}

We include additional qualitative rollouts to make the generated interactive worlds easier to inspect visually.
Figures~\ref{fig:margit_rollout_1min} and~\ref{fig:kof_rollout_0} complement the quantitative evaluation by showing representative long-horizon behavior in the two domains used in the paper.

% \subsection{Additional Qualitative Results}
% \label{app:visual_results}

% We further present additional qualitative results of \methodname
% operating on the game \textit{The King of Fighters (KOF)},
% as shown in \Cref{fig:kof_demo}.
% \methodname achieves fine-grained multi-entity action control
% and accurately captures highly brief actions
% (e.g., \textit{Punching}) lasting within $0.25$\,s.

% \begin{figure*}[t]
%     \centering
%     \captionsetup{justification=raggedright, singlelinecheck=false}
%     \includegraphics[width=\linewidth]{figures/fig_kof_demo.pdf}
%     \caption{\textbf{Demonstrations of fine-grained multi-entity action control of \methodname in \textit{KOF}.}
%     \methodname precisely responds to rapid action inputs and successfully captures
%     actions as brief as $0.25$\,s (e.g., \textit{Punching}),
%     demonstrating its fine-grained and responsive control capability.}
%     \label{fig:kof_demo}
% \end{figure*}

\subsection{Full Action Vocabulary}
\label{app:vocab}

We summarize the per-entity action vocabularies used for prompt conditioning in Table~\ref{tab:vocab}. The player vocabulary $\mathcal{A}_{\text{player}}$ consists of $13$ actions covering locomotion, defensive rolls, weapon attacks, and terminal states. Margit's native repertoire $\mathcal{A}_{\text{boss}}^{\text{Margit}}$ contains $30$ actions, while the Crucible Knight contributes $17$ additional non-overlapping moves. We obtain the joint boss vocabulary $\mathcal{A}_{\text{boss}}^{\text{joint}}$ with $|\mathcal{A}_{\text{boss}}^{\text{joint}}|=47$ by deduplication across the two bosses, and we adopt this joint vocabulary throughout the experiments so that any cross-entity action index is technically injectable into either boss's context.

We obtain these vocabularies by manually aggregating the raw animation-state IDs read from engine memory. The raw stream is not an action vocabulary in the human-meaningful sense: a single human action such as a heavy slash unrolls into a sequence of typically six or more consecutive raw IDs corresponding to its sub-phases (e.g., \texttt{windup}\,$\to$\,\texttt{strike}\,$\to$\,\texttt{recovery}\,$\to$\,\texttt{idle}), and a representative recording session already exposes $111$ distinct player IDs and $53$ distinct boss IDs even before the full dataset is exhausted. Domain-expert aggregation from the raw IDs to the $13$/$47$-action vocabulary is therefore a prerequisite shared by any Action-Index baseline rather than an advantage of NL conditioning, and our vocabularies define the action-level ground truth on which both NL and Action-Index baselines are evaluated. We release the full raw-ID-to-action mapping with the dataset (see Appendix~\ref{app:stronger_id} for the implications on stronger Action-Index baselines).

\begin{table}[h]
  \caption{Margit's native action vocabulary, with $|\mathcal{A}_{\text{player}}|=13$ and $|\mathcal{A}_{\text{boss}}^{\text{Margit}}|=30$. The Crucible Knight contributes additional non-overlapping actions, yielding the joint boss vocabulary $|\mathcal{A}_{\text{boss}}^{\text{joint}}|=47$ that we use throughout the experiments.}
  \label{tab:vocab}
  \centering
  \small
  \begin{tabular}{p{0.45\linewidth}p{0.45\linewidth}}
    \toprule
    \textbf{Player Actions ($\mathbf{13}$)} & \textbf{Boss Actions ($\mathbf{30}$)} \\
    \midrule
    Standing, Move ($4$dir), Roll ($4$dir), & Moving (forward\,/\,left\,/\,right), \\
    Greatsword Sweep, Greatsword Thrust, & Jump and mid-air slam, Horizontal slash, \\
    Death, Execution & Heavy overhead slash, Staff slam, \\
    & Uttering curse, Quick slam, \\
    & Double light blade throw, Staff upswing, \\
    & Tail swipe, Double light blade slash, \\
    & X-shaped slash, Charged staff thrust, \\
    & Forward charge, Jump back to disengage, \\
    & $+\,13$ additional combo/variant moves \\
    \bottomrule
  \end{tabular}
\end{table}

\subsection{Baseline Prompt Settings}
\label{app:baseline_prompt}

We specify the text prompts that we feed to the three video generation baselines compared in \Cref{fig:baseline_comparison}, namely
\textbf{Seedance~$\mathbf{2.0}$}~\citep{seedance2026seedance20advancingvideo},
\textbf{Kling~$\mathbf{3.0}$}~\citep{kuaishou2026kling}, and
\textbf{LongLive}~\citep{yang2025longlive}, as follows.

\begin{quote}
\noindent\textbf{Environment:} Stormveil Castle bridge, overcast sky, cinematic combat.\\
\textbf{Agents:} Player (Greatsword user) vs.\ Boss (Margit, the Fell Omen).
\noindent\textbf{Player:}
\begin{itemize}[noitemsep, topsep=2pt, leftmargin=1.5em]
    \item $0.00$\,s\,--\,$1.50$\,s:\; Move forward
    \item $1.50$\,s\,--\,$2.50$\,s:\; Roll forward
    \item $2.50$\,s\,--\,$3.50$\,s:\; Greatsword thrust
    \item $3.50$\,s\,--\,$4.50$\,s:\; Roll forward
    \item $4.50$\,s\,--\,$5.50$\,s:\; Greatsword thrust
    \item $5.50$\,s\,--\,$6.25$\,s:\; Roll backward
    \item $6.25$\,s\,--\,$7.25$\,s:\; Greatsword thrust
    \item $7.25$\,s\,--\,$8.00$\,s:\; Roll left
    \item $8.00$\,s\,--\,$8.75$\,s:\; Roll right
    \item $8.75$\,s\,--\,$10.00$\,s:\; Move forward
\end{itemize}
\noindent\textbf{Boss:}
\begin{itemize}[noitemsep, topsep=2pt, leftmargin=1.5em]
    \item $0.00$\,s\,--\,$2.50$\,s:\; Jump and mid-air slam
    \item $2.50$\,s\,--\,$4.00$\,s:\; Tail swipe
    \item $4.00$\,s\,--\,$5.00$\,s:\; Jump back to disengage
    \item $5.00$\,s\,--\,$7.50$\,s:\; Jump and mid-air slam
    \item $7.50$\,s\,--\,$9.00$\,s:\; Tail swipe
    \item $9.00$\,s\,--\,$10.00$\,s:\; Horizontal slash
\end{itemize}
\end{quote}

Since the three commercial baselines all incorporate built-in prompt-enhancement modules, we adopt this explicit timestamp-structured format to ensure a fair and controlled comparison with \methodname under matched per-entity action schedules.

\subsection{Experimental Setup Details}
\label{app:setup_details}
\label{app:details}

\paragraph{Why games are the testbed.}
We choose games as the testbed because the interface claim requires frame-accurate multi-entity action labels at interactive frame rates, and games are the only domain that offers all three properties simultaneously. Specifically, per-frame animation state is readable from engine memory at zero annotation cost, the action vocabularies are bounded yet non-trivial, and the evaluation criteria are unambiguous. In contrast, driving and embodied-manipulation datasets lack frame-level entity-wise annotations, and narrative-video datasets lack adversarial multi-entity dynamics.

\paragraph{Data pipeline.}
We assemble the training data from three distinct entity domains:
\begin{itemize}[leftmargin=*,topsep=2pt,itemsep=2pt]
  \item \textbf{Elden Ring -- Margit:} We collect $30$ hours of boss-fight gameplay and segment it into ${\sim}10{,}000$ high-quality $5$-second clips at $16$\,FPS after filtering and quality-based pruning, with $10\%$ held out by recording date for evaluation.
  \item \textbf{Elden Ring -- Crucible Knight:} We collect $15$ hours of comparable footage segmented and filtered identically (${\sim}5{,}000$ clips), and we use it jointly with Margit for the cross-entity evaluation.
  \item \textbf{The King of Fighters (KOF):} We collect ${\sim}5{,}000$ $60$-second fighter-pair clips at $16$\,FPS (${\approx}83$ hours in total), and we use this corpus to validate the cross-world transfer of the architecture.
\end{itemize}
For Elden Ring, we obtain all per-frame labels by reading the engine's \texttt{current\_animation} field at runtime via direct memory instrumentation, which yields zero-offset action triplets $(v_t, a_t^{\text{player}}, a_t^{\text{boss}})$ with $a_t \in \mathcal{A}_{\text{player}} \times \mathcal{A}_{\text{boss}}^{\text{joint}}$. We have $|\mathcal{A}_{\text{player}}|=13$ and $|\mathcal{A}_{\text{boss}}^{\text{joint}}|=47$, where the joint boss vocabulary is the deduplicated union of Margit and the Crucible Knight (Margit's native subset contains $30$ actions; see Appendix~\ref{app:vocab}). For KOF, we read per-frame labels from the emulator's animation-state register under the same zero-cost protocol.

\begin{figure*}[t]
    \centering
    \includegraphics[width=\linewidth]{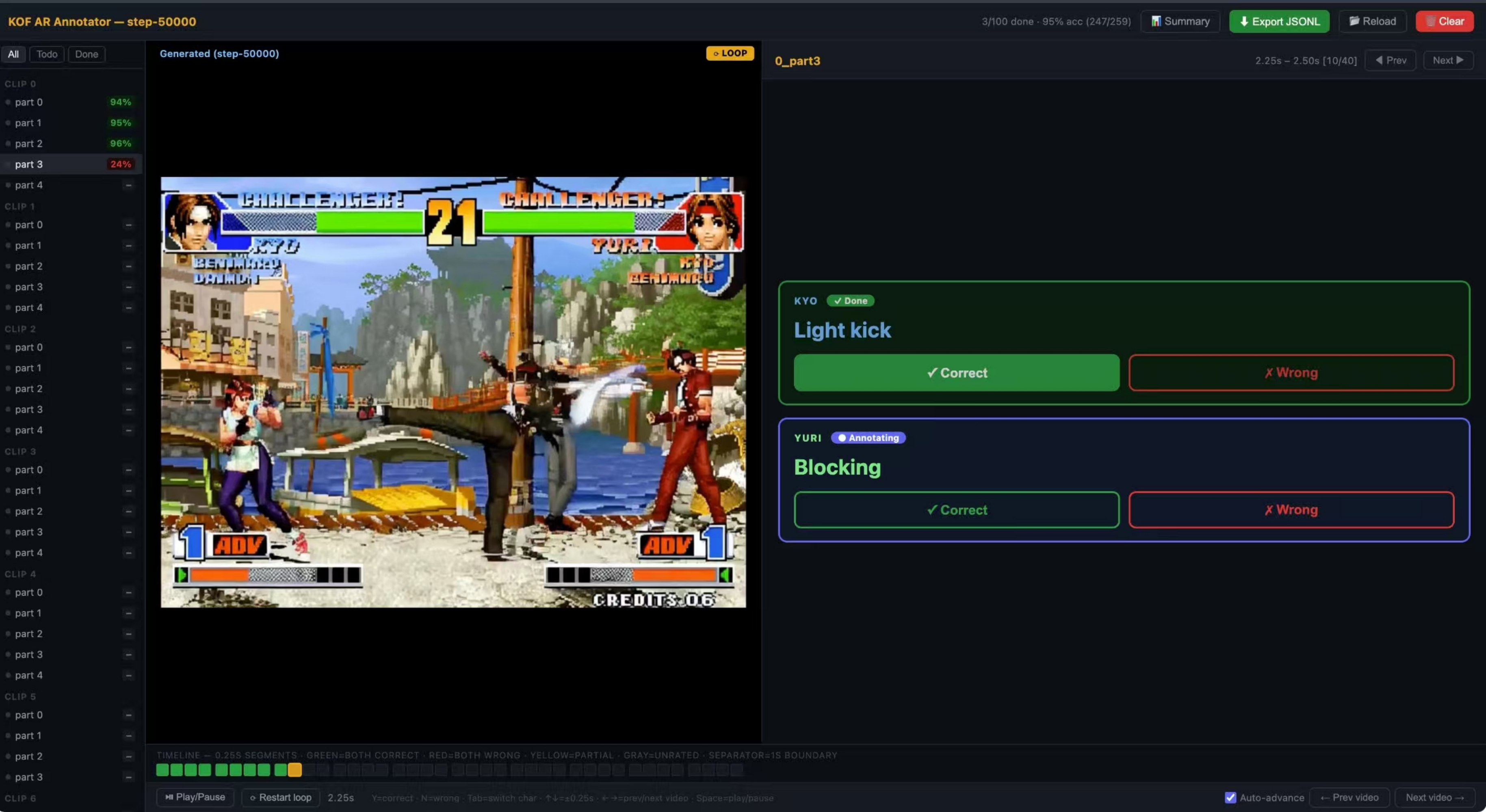}
    \caption{\textbf{Annotation interface for the human evaluation of Action Control Accuracy (ACA).}
    Each trial presents the annotators with a generated video clip alongside the per-entity target action label (here: \textsc{Kyo}---\textit{Light kick}; \textsc{Yuri}---\textit{Blocking}).
    We strip conditioning-variant identities (NL vs.\ Action-Index) and prompt-source labels before rating, which ensures a fully blinded evaluation.
    Each annotator rates each entity's action on the three-point ordinal scale ($\mathbf{0}$: absent; $\mathbf{1}$: partial; $\mathbf{2}$: full); we then take the per-clip ACA($\geq 1$) as the binary indicator that the median of the three ratings is at least $1$.}
    \label{fig:annotator}
\end{figure*}

\paragraph{Annotation protocol.}
For every reported ACA number, we pool all $20$ trials 
per condition ($5$ starting frames $\times$ $4$ seeds) 
across all conditions, randomly shuffle them, and have three 
annotators independently rate each clip on a three-point 
ordinal scale ($\mathbf{0}$: action absent; $\mathbf{1}$: 
partial execution; $\mathbf{2}$: full execution). 
Before rating, we strip both the conditioning-variant label 
(NL vs.\ Action-Index) and the prompt-source identity 
(in-distribution vs.\ cross-entity vs.\ OOV), so that 
all clips are rated under fully blinded conditions,
as shown in the annotation interface in \Cref{fig:annotator}. 
We take the per-trial score as the median of the three ratings, and we report ACA($\geq 1$) as the fraction of clips whose median is at least $1$.

\paragraph{Prompt-injection rollout protocol (Axes 1--3).}
We adopt the following prompt-injection protocol for all interface-evaluation rollouts in Axes 1--3, where the goal is to isolate the model's compositional steering capability:
\begin{enumerate}[leftmargin=*,topsep=2pt,itemsep=2pt,label=(\roman*)]
  \item \textbf{Starting frame.} We sample a starting frame uniformly at random from the held-out test split and decode it into the model's visual context window.
  \item \textbf{Un-conditioned warm-up.} The model then generates approximately $2$ seconds ($32$ frames) of un-conditioned video, during which we set both the player and boss action prompts to the neutral idle/standing token at every step. The model is therefore conditioned only on its own visual history. This warm-up serves two purposes. First, it places the model in a steady-state denoising regime before the prompted phase begins, which avoids the artifacts typical of cold-started rollouts. Second, it severs any residual visual cue from the test-set continuation that might otherwise inform the model that a particular action is about to occur; without this, the model could in principle reproduce the target action by extrapolating the test-set trajectory rather than by genuinely responding to the prompt.
  \item \textbf{Prompt injection.} At $t=2$\,s, we replace the neutral prompt with the target action prompt and hold it constant for the remaining $3$ seconds of the clip.
  \item \textbf{Rating.} Annotators rate the resulting $5$-second clip against the target action over the post-injection window.
\end{enumerate}
We apply the same warm-up and injection schedule to both the NL and Action-Index conditioning variants under matched starting frames and seeds, so that the interface comparison is conducted under identical visual conditions. We do not search over the warm-up duration: the $2$-second value is fixed before annotation begins.

\paragraph{Trajectory-conditioned protocol (system metrics).}
We adopt a separate trajectory-conditioned protocol for Table~\ref{tab:system} and the system-level ablations in Appendices~\ref{app:stage1_ablation} and~\ref{app:stage2_ablation}, where the goal is to measure how faithfully the model tracks a fully specified ground-truth action trajectory rather than its compositional steering ability:
\begin{itemize}[leftmargin=*,topsep=2pt,itemsep=2pt]
  \item \emph{Sample.} We use $100$ held-out $10$-second clips per model.
  \item \emph{Conditioning.} Rollouts begin at the first frame of each clip, and at every frame we set both the player and boss prompts to the ground-truth caption derived from engine memory. We use no warm-up phase and no prompt switch.
  \item \emph{Rating.} We rate each rollout as binary correct or incorrect against its source clip's action sequence, and ACA reports the fraction correct.
\end{itemize}
The two protocols therefore evaluate complementary capabilities, namely compositional steering versus trajectory fidelity, and their absolute ACA values should not be directly compared.

\paragraph{Training.}
We fine-tune the Stage~1 teacher for $51$k iterations ($1$k warmup at $256\times448$ followed by $50$k at $480\times832$) using AdamW with peak learning rate $1\times10^{-5}$ and global batch size $64$ on $16\times$H$100$ $80$\,GB GPUs. We then perform ODE initialization for $1{,}000$ steps at $480\times832$ (learning rate $5\times10^{-6}$, batch size $128$, $16\times$H$100$), followed by Self-Forcing distillation for $15$k iterations at learning rate $2\times10^{-6}$.

\paragraph{Video resolution.}
We generate all videos at $480\times832$ ($480$p) and $16$\,FPS. We compress context frames into latents through the base model's VAE at a $4\times$ spatial downsampling and a $4\times$ temporal downsampling.

\paragraph{Hit detection.}
We fine-tune Qwen3-VL-$2$B-Instruct~\citep{bai2025qwen3} with LoRA on the vision--language connector and the cross-attention modules. We aggregate frame-level predictions into a per-window classification through majority voting.

\paragraph{End-to-end throughput.}
\label{app:throughput}
We measure the diffusion student's wall-clock latency at $160$\,ms per frame on a single H100 $80$\,GB, where the per-frame loop is dominated by the diffusion pass with KV-cache sliding and RoPE decoupling, the VAE decode, and the surrounding I/O. We will release detailed per-component timings alongside the open-source code.

\subsection{Annotator Reliability}
\label{app:reliability}

We rate $400$ generated clips in total ($200$ 
cross-entity and $200$ in-distribution) under the protocol 
of Appendix~\ref{app:setup_details}. Each clip is scored 
independently by three blinded annotators on the 
$\{0, 1, 2\}$ ordinal scale with conditioning-variant and 
prompt-source labels stripped, and we take the per-clip ACA 
score as the binary indicator that the median of the three 
ordinal ratings is at least $1$. In this subsection we 
report inter-rater consistency restricted to the $5$-pair 
cross-entity subset (the Axis~$2$ pairs in 
Table~\ref{tab:main_results}) and the $5$-action 
in-distribution subset (the Axis~$1$ actions in 
Table~\ref{tab:main_results}). 
% We exclude one candidate 
% action (m\_HorizontalSlash) from both subsets on 
% evaluation-side grounds: 
% the three-rater within-$1$ ordinal agreement on this 
% action drops to $75\%$, in clear contrast to 
% the $93$--$96\%$ range observed for all retained actions. 
% The exclusion is therefore driven by annotator reliability 
% rather than by interface behavior.

\paragraph{Within-$\mathbf{1}$ ordinal agreement.}
On the $\{0,1,2\}$ scale, raters disagree by more than one tier on fewer than $7\%$ of clips across either split. Specifically, the within-$1$ agreement on the cross-entity subset is $96.5\%$, $96.0\%$, and $95.0\%$ for the three rater pairs $A_1\!\times\!A_2$, $A_1\!\times\!A_3$, and $A_2\!\times\!A_3$; on the in-distribution subset, the corresponding numbers are $93.3\%$, $95.8\%$, and $94.2\%$. Consequently, aggregating by the median of the three ratings before binarizing at the $\geq 1$ threshold inherits a noise floor of at most one-tier disagreement on at most $7\%$ of clips, which is materially smaller than every NL\,vs.\,Action-Index gap reported in the paper.

\paragraph{Paired McNemar on cross-entity: $\mathbf{Z=6.38}$, $\mathbf{p<10^{-10}}$.}
For each (pair, starting frame, seed) triple, 
we obtain one NL clip and one Action-Index clip 
evaluated under identical visual conditioning and 
rated by the same three blinded annotators under the 
median-of-three scheme. We pool the matched triples 
across all five pairs and apply the McNemar test on the 
resulting binary scores. The test yields $49$ triples 
on which NL succeeds and Action-Index fails, against $3$ 
triples on which Action-Index succeeds and NL fails, 
giving $Z=6.38$ and $p<10^{-10}$ (one-sided). Therefore, among the clips on which the two interfaces produce different outcomes, the language interface succeeds on more than an order of magnitude as many clips as the Action-Index baseline, and the cross-entity gap cannot be attributed to cell-level fluctuation.

\subsection{Stage 1: Conditioning Architecture Ablation}
\label{app:stage1_ablation}

We ablate the two key Stage~$1$ design choices in a $2\times 2$ factorial that crosses the type of history self-attention (bidirectional vs.\ causal) with the scope of text cross-attention (noisy frame only vs.\ all frames). We train all four variants from the same Wan~$2.2$ TI2V-5B checkpoint with identical hyperparameters (batch size $64$, learning rate $2\times 10^{-5}$, resolution $256\times 448$), and we evaluate them on the held-out test split every $5$k steps up to $35$k steps under the trajectory-conditioned protocol of Appendix~\ref{app:setup_details}.

\begin{table}[h]
  \caption{\textbf{Stage~$\mathbf{1}$ architecture ablation ($\mathbf{2\times 2}$).} Each cell reports FVD$\downarrow$ at the best checkpoint, selected by the lowest FVD. $\dagger$: training never stabilizes, with FVD exceeding $1{,}100$ at all checkpoints.}
  \label{tab:ablation}
  \centering
  \setlength{\tabcolsep}{12pt}
  \begin{tabular}{lcc}
    \toprule
    & \textbf{Text: noisy only} & \textbf{Text: all frames} \\
    \midrule
    \textbf{Bidir.\ history (ours)} & $\mathbf{201.9}$ & $197.1$ \\
    \textbf{Causal history}         & $245.1$          & $1157.9^\dagger$ \\
    \bottomrule
  \end{tabular}
\end{table}

\begin{figure}[h]
\centering
\begin{tikzpicture}[font=\small]
\begin{axis}[
  width=0.92\linewidth,
  height=5.2cm,
  xlabel={Training Steps (k)},
  ylabel={FVD $\downarrow$},
  xmin=3, xmax=37,
  ymin=100, ymax=1800,
  xtick={5,10,15,20,25,30,35},
  xticklabels={5k,10k,15k,20k,25k,30k,35k},
  ytick={200,400,600,800,1000,1200,1400,1600,1800},
  ymajorgrids=true,
  grid style={gray!20, dashed},
  legend pos=north east,
  legend style={font=\scriptsize, fill=white, fill opacity=0.9,
                draw=gray!40, inner sep=3pt, row sep=1pt},
  tick label style={font=\scriptsize},
  label style={font=\scriptsize},
  clip=true,
]
\addplot[color=blue!80!black, thick, mark=*, mark size=2pt]
  coordinates {(5,353.1)(10,345.7)(15,252.3)(20,225.7)(25,275.7)(30,201.9)(35,233.1)};
\addlegendentry{Bidir + text on noisy \textbf{(ours)}}
\addplot[color=green!60!black, thick, mark=square*, mark size=2pt]
  coordinates {(5,325.4)(10,349.4)(15,248.9)(20,203.4)(25,204.5)(30,197.1)(35,223.2)};
\addlegendentry{Bidir + text on all frames}
\addplot[color=orange!80!black, thick, mark=triangle*, mark size=2.5pt, dashed]
  coordinates {(5,448.9)(10,427.2)(15,272.7)(20,255.8)(25,301.5)(30,245.1)(35,273.3)};
\addlegendentry{Causal + text on noisy}
\addplot[color=red!70!black, thick, mark=x, mark size=3pt, dashed]
  coordinates {(5,1386.4)(10,1644.9)(15,1500.7)(20,1430.4)(25,1157.9)(30,1173.6)(35,1653.1)};
\addlegendentry{Causal + text on all frames $\dagger$}
\node[font=\scriptsize, red!70!black, anchor=west]
  at (axis cs: 25.5, 1090) {temporal contamination};
\end{axis}
\end{tikzpicture}
\caption{\textbf{Stage~$\mathbf{1}$ ablation: FVD vs.\ training steps.} Companion to Table~\ref{tab:ablation}. The causal-history plus full-cross-attention configuration ($\dagger$) collapses throughout training, whereas the other three configurations converge to FVD$\,\sim 200$.}
\label{fig:ablation_curve}
\end{figure}
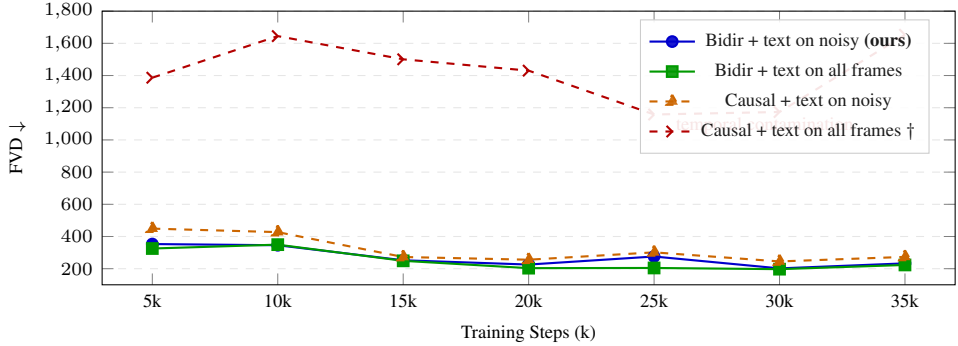

Three findings emerge from this ablation.
\textbf{($\mathbf{1}$) Bidirectional history attention consistently outperforms causal history attention} ($201.9$ vs.\ $245.1$ FVD). This confirms that forcing causal masking on history tokens breaks the bidirectional inductive bias inherited from the Wan~$2.2$ pretrained weights.
\textbf{($\mathbf{2}$) Decoupling text cross-attention to the noisy frame is essentially free} ($201.9$ vs.\ $197.1$ FVD, within noise), which demonstrates that our design preserves semantic clarity at zero quality cost.
\textbf{($\mathbf{3}$) Causal history combined with full cross-attention is unstable} (FVD $>1{,}100$ throughout training). In this configuration, injecting the current-frame action label $a_t$ into committed causal history keys contaminates those representations with future action information, which supports the temporal cross-contamination analysis in Section~\ref{sec:stage1}.
We further plot the per-step training curves in Figure~\ref{fig:ablation_curve}, and they confirm that these findings hold throughout training rather than only at a single checkpoint.

\subsection{Stage 2: KV-Cache and Bounded RoPE Ablation}
\label{app:stage2_ablation}

We sweep the two Stage~$2$ 
design choices on the same 
distilled student. 
Group~A omits the bounded sliding window 
entirely and retains the full history, which incurs unbounded VRAM growth. Group~B fixes the sliding window at \texttt{kv\_window}$=7$ to match the training value $K_r=7$, and we sweep the local-RoPE cap within this group. Group~C fixes the cap at \texttt{cap}$=16$ and we sweep the window size. We report the held-out FVD under the trajectory-conditioned protocol of Appendix~\ref{app:setup_details} at both $10$\,s and $30$\,s horizons.

\begin{table}[h]
  \caption{\textbf{Stage~$\mathbf{2}$: KV-cache and RoPE ablation.} We report FVD$\downarrow$ on the $10$\,s and $30$\,s held-out test sets. $\ddagger$: full history retained in VRAM, with OOM risk on long sequences. KV sliding is the dominant factor: without it, FVD more than doubles at $30$\,s. With sliding enabled, the RoPE cap has negligible effect on FVD because the sliding window already bounds the 
  relative positions to the training range. All results are obtained using the native Wan $2.2$ VAE.}
  \label{tab:kvcache}
  \centering
  \setlength{\tabcolsep}{10pt}
  \begin{tabular}{llcc}
    \toprule
    Group & Configuration & FVD $\downarrow$ ($10$\,s) & FVD $\downarrow$ ($30$\,s) \\
    \midrule
    A: no sliding$^\ddagger$ & No cap                                 & $439.6$          & $996.9$          \\
    \midrule
    \multirow{2}{*}{B: kv$=7$}
                             & No cap                                 & $137.7$          & $141.2$          \\
                             & \textbf{cap$\mathbf{=16}$ (ours)}      & $\mathbf{138.6}$ & $\mathbf{139.2}$ \\
    \midrule
    C: cap$=16$              & kv$=4$                                 & $140.5$          & $141.9$          \\
    \bottomrule
  \end{tabular}
\end{table}

We summarize three findings.
\textbf{($\mathbf{1}$) KV sliding dominates, with benefits compounding at long horizon.}
Without sliding (Group~A), FVD rises from $439.6$ at $10$\,s to $996.9$ at $30$\,s, a $2.3\times$ degradation as the unbounded history strains memory and accumulates rollout errors. With sliding enabled (Groups~B and~C), FVD remains essentially flat across both horizons, with at most $3$ FVD points of change.
\textbf{($\mathbf{2}$) The RoPE cap has negligible empirical effect under sliding.}
The no-cap and cap$=16$ rows of Group~B differ by only $0.9$ FVD at $10$\,s and $2.0$ 
FVD at $30$\,s. The reason is that, 
under a $K_r = 7$ sliding window, the relative distance between the noisy target and any recent 
frame is bounded to $[1, 7]$ exactly by construction, 
which already covers the training range for target--recent attention.
The local-RoPE cap $C$ instead bounds the relative distance from the target to the sink frame, 
which would otherwise grow unboundedly as $p_t^{\mathrm{abs}}$ advances.
Empirically, the no-cap baseline still performs well because attention mass on the sink is small 
(the sink primarily encodes static scene anchors), so the OOD positional regime there has 
limited effect on FVD.
\textbf{($\mathbf{3}$) Window size $\mathbf{K_r=7}$ aligns with training.}
With \texttt{kv}$=4$ (Group~C), FVD stays within $3$ points of the \texttt{kv}$=7$ baseline at both horizons ($140.5$ vs.\ $138.6$ at $10$\,s; $141.9$ vs.\ $139.2$ at $30$\,s). This confirms that, once the relative-position bound is in place, the exact window length exerts only a second-order effect on quality.

\subsection{Stronger Action-Index Baselines Collapse into NL}
\label{app:stronger_id}

A natural intermediate baseline that might appear stronger than our hash-projected Action-Index interface is one whose embeddings are factorized into separate entity and action tables and initialized from a frozen text encoder such as CLIP, Qwen, or Wan's own. We argue that any such baseline is, by construction, a vocabulary-restricted special case of NL conditioning and therefore does not constitute a separate operating point. We analyze its two natural instantiations as follows.
\textbf{Frozen variant.} When we keep the text-initialized embeddings frozen, the baseline inherits exactly the pretrained semantic prior that NL exploits, and we therefore expect it to recover most of the Tier~II and Tier~III cross-entity gap. However, its inference vocabulary remains closed, so its OOV coverage on the four probes of Section~\ref{sec:language_vs_id} stays exactly $\mathbf{0}\%$. No choice of initialization can change a structural slot count.
\textbf{Trainable variant.} When we unfreeze the embeddings, the text-initialized entries specialize to their training-time entity context within a few thousand steps, and the variant is empirically dominated by the joint-vocabulary Action-Index baseline reported in Section~\ref{sec:language_vs_id}.
In either case, a factorized text-initialized Action-Index interface either collapses into NL with a hand-fixed sub-vocabulary (frozen) or reverts to the joint-vocabulary baseline (trainable).

\subsection{Axis 1: Full In-Distribution Score Distributions}
\label{app:full_indist}

We extend the main-body Axis~$1$ summary (Table~\ref{tab:main_results}) with the per-trial $2/1/0$ score distributions for both interfaces in Table~\ref{tab:indist_full}. We evaluate both NL and Action-Index on the same $20$ trials per action ($5$ starting frames $\times$ $4$ seeds), with three blinded annotators and median rating. We select the five evaluated actions as the most frequent in-distribution actions per entity, since these dominate each entity's training data and therefore provide the strongest possible supervision for the Action-Index interface, ruling out long-tail artifacts as an explanation for any subsequent NL\,vs.\,Action-Index gap.

\begin{table}[h]
  \caption{\textbf{Axis~$\mathbf{1}$ full score distributions.} Each cell reports the percentage of trials rated $\mathbf{2}$ (full execution), $\mathbf{1}$ (partial), and $\mathbf{0}$ (absent), with ACA($\geq 1$) defined as the sum of the full and partial percentages.}
  \label{tab:indist_full}
  \centering
  \setlength{\tabcolsep}{4pt}
  \begin{tabular}{lcccc}
    \toprule
    & \multicolumn{2}{c}{\textbf{NL (ours)}} & \multicolumn{2}{c}{\textbf{Action-Index}} \\
    \cmidrule(lr){2-3}\cmidrule(lr){4-5}
    Action
      & $2\,/\,1\,/\,0$\,(\%) & ACA($\geq 1$)
      & $2\,/\,1\,/\,0$\,(\%) & ACA($\geq 1$) \\
    \midrule
    Margit \textperiodcentered{} Double light blade throw
      & $60\,/\,35\,/\,\phantom{0}5$ & $\mathbf{95}$
      & $45\,/\,40\,/\,15$           & $85$            \\
    Margit \textperiodcentered{} Staff slam
      & $70\,/\,10\,/\,20$           & $\mathbf{80}$
      & $50\,/\,30\,/\,20$           & $\mathbf{80}$   \\
    Knight \textperiodcentered{} Shield block
      & $95\,/\,\phantom{0}5\,/\,\phantom{0}0$ & $\mathbf{100}$
      & $65\,/\,35\,/\,\phantom{0}0$           & $\mathbf{100}$ \\
    Knight \textperiodcentered{} Overhead slash
      & $100\,/\,\phantom{0}0\,/\,\phantom{0}0$ & $\mathbf{100}$
      & $75\,/\,10\,/\,15$                      & $85$           \\
    Knight \textperiodcentered{} Tail of the crucible
      & $70\,/\,30\,/\,\phantom{0}0$           & $\mathbf{100}$
      & $50\,/\,45\,/\,\phantom{0}5$           & $95$            \\
    \midrule
    \textbf{Mean ($\mathbf{5}$ actions)} & & $\mathbf{95}$ & & $89$ \\
    \bottomrule
  \end{tabular}
\end{table}

\subsection{Axis 2: Full Cross-Entity Distributions and Per-Tier Mechanism}
\label{app:axis2_tiers}

We extend the main-body Axis~$2$ summary (Table~\ref{tab:main_results}) with the per-trial $2/1/0$ score distributions in Table~\ref{tab:cross_entity_full}. We evaluate both interfaces on the same five cross-entity action pairs under identical annotation ($20$ trials per pair, $5$ starting frames $\times$ $4$ seeds, three blinded annotators with median rating, prompt-injection protocol of Appendix~\ref{app:setup_details}). In the rest of this subsection we first define the three tiers used to organize the pairs, and then we decompose the cross-entity NL\,vs.\,Action-Index gap by tier.

\paragraph{Definition of the three tiers.}
We grade each cross-entity action pair (source action, target entity) by the relationship between the source action and the \emph{target entity's} native action repertoire, since this relationship determines what the Action-Index interface can possibly fall back on when an out-of-context index is injected. We adopt three tiers of increasing visual overlap:
\begin{itemize}[leftmargin=*,topsep=2pt,itemsep=2pt]
  \item \textbf{Tier~$\mathbf{I}$ (no overlap):} The source action is entirely absent from the target entity's native repertoire, with no morphologically related fallback animation available. For example, the ``double light blade throw'' is exclusive to Margit and has no counterpart in the Crucible Knight's repertoire. This is the most stringent regime for the Action-Index interface, since its index-to-embedding map has no nearest neighbor to recover.
  \item \textbf{Tier~$\mathbf{II}$ (motion shared, visual style distinct):} The target entity possesses an action that shares the underlying motion class (e.g., a tail swipe) with the source action, but the two animations differ in a visually salient feature such as luminosity or trajectory shape. For instance, both Margit and the Knight execute a tail swipe, yet only the Knight's variant emits a luminous energy trail. Action-Index can fall back on the shared motion class, but it cannot produce the entity-specific visual feature.
  \item \textbf{Tier~$\mathbf{III}$ (same action label, animation alignment varies):} Both entities possess an action carrying the same lexical label (slash, overhead, horizontal), yet the underlying animations may be more or less tightly aligned in timing and amplitude. This is in principle the easiest regime for Action-Index, since its embedding can in principle map onto a visually adjacent animation.
\end{itemize}
We choose this stratification so that the three tiers progressively give the Action-Index interface more and more chance to succeed via nearest-neighbor fallback. If the cross-entity NL advantage persists across all three tiers, then the gap cannot be attributed to a single failure mode of the Action-Index interface.

\begin{table}[h]
  \caption{\textbf{Axis~$\mathbf{2}$ full score distributions.} Each cell reports the percentage of trials rated $\mathbf{2}$ (full) / $\mathbf{1}$ (partial) / $\mathbf{0}$ (absent), with ACA($\geq 1$) defined as the sum of the full and partial percentages. We grade tiers by the degree of visual overlap between the source action and the target entity's native repertoire.}
  \label{tab:cross_entity_full}
  \centering
  \setlength{\tabcolsep}{3pt}
  \begin{tabular}{llccccc}
    \toprule
    &     & \multicolumn{2}{c}{\textbf{NL (ours)}} & \multicolumn{2}{c}{\textbf{Action-Index}} \\
    \cmidrule(lr){3-4}\cmidrule(lr){5-6}
    Action prompt (source entity) & Target entity
      & $2\,/\,1\,/\,0$\,(\%) & ACA($\geq 1$)
      & $2\,/\,1\,/\,0$\,(\%) & ACA($\geq 1$) \\
    \midrule
    \multicolumn{6}{l}{\textit{Tier~$\mathbf{I}$ --- action absent from the target entity's native repertoire}} \\
    \quad Double light blade throw (Margit) & Knight
      & $65\,/\,15\,/\,20$           & $\mathbf{80}$
      & $\phantom{0}0\,/\,15\,/\,85$ & $15$           \\
    \midrule
    \multicolumn{6}{l}{\textit{Tier~$\mathbf{II}$ --- target has a morphologically similar but visually distinct action}} \\
    \quad Tail of the crucible (Knight) & Margit
      & $20\,/\,70\,/\,10$           & $\mathbf{90}$
      & $\phantom{0}0\,/\,60\,/\,40$ & $60$           \\
    \midrule
    \multicolumn{6}{l}{\textit{Tier~$\mathbf{III}$ --- both entities share a same-named action; Action-Index may fall back to it}} \\
    \quad Heavy overhead slash (Margit) & Knight
      & $95\,/\,\phantom{0}5\,/\,\phantom{0}0$ & $\mathbf{100}$
      & $60\,/\,15\,/\,25$                     & $75$           \\
    \quad Diagonal slash (Knight)       & Margit
      & $45\,/\,50\,/\,\phantom{0}5$           & $\mathbf{95}$
      & $\phantom{0}5\,/\,20\,/\,75$           & $25$           \\
    \quad Horizontal slash (Margit)     & Knight
      & $70\,/\,10\,/\,20$                     & $\mathbf{80}$
      & $\phantom{0}0\,/\,40\,/\,60$           & $40$           \\
    \midrule
    \textbf{Mean ($\mathbf{5}$ pairs)} & & & $\mathbf{89}$ & & $43$ \\
    \bottomrule
  \end{tabular}
\end{table}

\paragraph{Per-tier decomposition of the gap.}
We now expand on the cross-entity result of Section~\ref{sec:language_vs_id} by tier. We argue that the three tiers of Table~\ref{tab:cross_entity_full} together rule out every single-confounder explanation of the cross-entity gap.

\textbf{Tier~$\mathbf{I}$ (NL$=80\%$, Action-Index$=15\%$).}
We observe the largest single-pair gap of $+65$\,pp here. The Crucible Knight is never paired with the ``double light blade throw'' index during training, so this tier is also the most stringent point of the prompt-injection protocol: the cross-entity action has zero training co-occurrence with the target entity, and the un-conditioned warm-up therefore anchors the model in an arbitrary Knight pose with no preparatory frames for the requested throw, which places visual coherence and semantic compliance maximally in tension. Under this regime, the Action-Index baseline reaches only $15\%$ ACA, with all successes registering as partial rather than full execution ($0\%$ rating-$2$, $15\%$ rating-$1$). We interpret this score-distribution pattern as the intended diagnostic signature of the protocol and as consistent with an embedding-leakage account: the Margit-trained ``double light blade'' embedding does carry enough action-specific visual gradient that, when injected into Knight context, the backbone occasionally synthesizes partial luminous-blade content; however, the Knight's vocabulary contains no nearest-neighbor action that visually resembles a thrown pair of blades, so the embedding has no usable fallback, and the partial visual content fails to crystallize into a full execution. NL, in contrast, reaches $80\%$ on the same prompt, with $65\%$ rating-$2$ full executions and $15\%$ rating-$1$ partials. The reason is that ``double light blade throw'' decomposes into subword units that the text encoder has seen in many training contexts (light, blade, throw), and the compositional meaning therefore transfers to the Knight without requiring any (Knight, light-blade) training co-occurrence. The language interface thereby largely resolves the visual-coherence-versus-semantic-compliance tension that the prompt-injection protocol creates by construction. We further note that, when the same student is evaluated under the trajectory-conditioned protocol used for the system-level metrics in Table~\ref{tab:system}, where ground-truth captions accompany the visual context from frame zero and the visual--semantic tension is removed, ACA reaches $90.4$--$93.2\%$, which is the regime in which the model is actually deployed during real-time play.

\textbf{Tier~$\mathbf{II}$ (NL$=90\%$, Action-Index$=60\%$).}
We observe a $+30$\,pp gap on this tier. The source action is the Crucible Knight's luminous energy tail, while Margit possesses a structurally analogous tail-swipe action with a dark, opaque visual style. We attribute the relatively high non-zero Action-Index ACA ($60\%$) to nearest-neighbor fallback: the injected index activates Margit's existing tail motion. However, Action-Index cannot encode the luminous visual feature, since this feature is specific to the Knight's animation. NL reaches $90\%$, with $70\%$ of clips judged as partial rather than full execution. We attribute this to the fact that the prompt ``Tail of the crucible'' encodes both the motion and the luminous visual characteristic through its pretrained semantics, which suppresses Margit's native dark-tail prior. The high partial-execution rate suggests, however, that the luminous quality is rendered less crisply on Margit than on its native Knight animation in Tier~III shared-action cases.

\textbf{Tier~$\mathbf{III}$ (Action-Index$\in [25\%, 75\%]$).}
We observe that, when the source action has a same-named counterpart in the target entity's native repertoire (slash, overhead, horizontal), the Action-Index nearest-neighbor fallback is in principle the strongest, because the index-to-embedding map should land on a visually adjacent animation. Empirically, however, the fallback is sharply contingent on \emph{animation alignment} rather than on the shared label.
We illustrate this through three pairs of increasing animation mismatch.
\textit{Heavy overhead slash} is the most aligned pair, since Margit's and the Knight's overhead slashes share both timing and amplitude; Action-Index therefore reaches $75\%$ ACA, the highest cross-entity Action-Index number in the table.
\textit{Horizontal slash} carries an identical lexical label, but the two underlying animations differ in sweep duration; Action-Index drops to $40\%$.
\textit{Diagonal slash} is the most extreme case: the Knight's variant is a short jab while Margit's is a wide arcing swing, so the shared index lands on visually mismatched footage, and Action-Index falls to $25\%$. We highlight that this Tier~III value ($25\%$) is even \emph{below} the Tier~II Action-Index number ($60\%$), where at least the underlying motion class is shared. The variation across Tier~III is therefore governed by animation-level visual specificity rather than by the lexical fact that the two entities share an action name. NL, in contrast, stays above $80\%$ on all three pairs, because language conditions on motion semantics independently of which animation index the target entity happens to own. We further note the residual NL\,$\geq$\,Action-Index gap on the easiest possible Action-Index case (Heavy overhead slash, $+25$\,pp), which indicates that language provides richer steering even on unambiguous shared actions, rather than merely compensating for missing vocabulary entries.

\paragraph{Summary.}
The three tiers together rule out every single-confounder explanation of the cross-entity gap. The NL advantage is not merely that ``Action-Index has no embedding for the action'' (Tier~$\mathbf{I}$, $+65$\,pp), nor merely that ``Action-Index picks the wrong visual style'' (Tier~$\mathbf{II}$, $+30$\,pp); rather, it persists even when Action-Index has both the right embedding and the right visual style (Tier~$\mathbf{III}$, Heavy overhead slash, $+25$\,pp). We therefore conclude that the gap tracks the structural property that NL possesses and Action-Index lacks, namely the compositional decomposition of action prompts into entity-independent semantics, rather than any single failure mode of the index-bound interface.

% ather than any single failure mode of the index-bound interface.

\paragraph{VLM pairwise corroboration.}
To provide an automated sanity check complementary to the human annotation, we run a vision-language model (VLM) judge on the same five cross-entity pairs.
For each pair we uniformly sample $10$ frames from the action burst segment of each video and present all $20$ frames to \texttt{gemini-3.1-flash-lite-preview} (accessed via API), randomising which video is labelled A and which is B.
The model is prompted with the action name, a concise visual definition, and asked to (i) score each video on a $0$--$3$ action-fidelity scale and (ii) declare a winner (\textit{A}, \textit{B}, or \textit{tie}).
We repeat this for $n{=}20$ matched NL/Action-Index pairs per action and report the fraction of trials won by each interface (Table~\ref{tab:vlm_pairwise}).

\begin{table}[h]
  \caption{\textbf{VLM pairwise judge ($10$ frames per video, $n{=}20$ pairs, \texttt{gemini-3.1-flash-lite-preview}).}
    Each trial presents both videos with randomised A/B assignment; the model chooses which clip better executes the named action.
    NL win\% and ID win\% are the fractions of trials won by each interface; ties account for the remainder.
    $\Delta = \text{NL win\%} - \text{ID win\%}$.}
  \label{tab:vlm_pairwise}
  \centering
  \setlength{\tabcolsep}{6pt}
  \begin{tabular}{llccc}
    \toprule
    Tier & Action & NL win\% & ID win\% & $\Delta$ \\
    \midrule
    I   & Double light blade   & $85$ & $15$ & $+70$\,pp \\
    II  & Tail of the crucible & $50$ & $50$ & ${\approx}0$\,pp \\
    III & Heavy overhead slash & $60$ & $40$ & $+20$\,pp \\
    III & Diagonal slash       & $55$ & $45$ & $+10$\,pp \\
    III & Horizontal slash     & $60$ & $35$ & $+25$\,pp \\
    \bottomrule
  \end{tabular}
\end{table}

The VLM results largely corroborate the human annotations.
On Tier~I, the model assigns NL a $+70$\,pp advantage, consistent with the $+65$\,pp human gap and confirming the most unambiguous transfer case.
On Tier~III, NL leads by $+10$--$+25$\,pp across all three pairs, matching the human-observed pattern that Action-Index can partially fall back on a same-named animation but cannot match the steering precision of language.
The sole divergence is Tier~II (\textit{Tail of the crucible}): the VLM produces chance-level agreement ($50\%$/$50\%$), because the luminous energy tail is an out-of-distribution visual element on Margit---the VLM has no game-domain prior to distinguish it from her native golden-blade effects, and therefore cannot reliably judge which clip is correct.
This reflects an inherent limitation of general-purpose VLM judges on domain-specific cross-entity transfers: when the target action is visually OOD for the target entity, only human annotators with game context can evaluate it reliably.

\subsection{OOV Probe Set}
\label{app:oov_probes}

We evaluate four out-of-vocabulary (OOV) probes referenced in Section~\ref{sec:language_vs_id}, each obtained by editing exactly one content word of a top-$3$-by-frequency in-vocabulary base prompt. We run each probe for $10$ trials ($5$ starting frames $\times$ $2$ seeds) under the prompt-injection protocol of Appendix~\ref{app:setup_details}, and we have two annotators rate each clip on the same ordinal scale used elsewhere; ACA($\geq 1$) reports the sum of correct and partial ratings. By construction, none of the four probes corresponds to an index in the $47$-way joint vocabulary, so the Action-Index interface returns no output for any of them and its coverage is structurally $\mathbf{0}\%$, regardless of model capacity.

\begin{table}[h]
  \caption{\textbf{OOV probe results (NL).} Per-probe ACA on the four-probe evaluation set, with aggregate ACA of $\mathbf{90}\%$ over $40$ trials. Action-Index coverage on each probe is structurally $\mathbf{0}\%$.}
  \label{tab:oov_probes}
  \centering
  \setlength{\tabcolsep}{4pt}
  \begin{tabular}{llcc}
    \toprule
    Base action (entity)                & OOV prompt (edit type)                       & Trials & ACA($\geq 1$) \\
    \midrule
    Overhead slash (Knight)             & \texttt{Downward slash.} (synonym)           & $10$   & $100$            \\
    Tail of the crucible (Knight)       & \texttt{Crucible tail.} (abbreviation)       & $10$   & $100$            \\
    Shield block (Knight)               & \texttt{Shield guard.} (synonym)             & $10$   & $\phantom{0}90$  \\
    Double light blade throw (Margit)   & \texttt{Dual light blade throw.} (synonym)   & $10$   & $\phantom{0}70$  \\
    \midrule
    \textbf{Mean ($\mathbf{4}$ probes)} &                                              & $40$   & $\mathbf{90}$    \\
    \bottomrule
  \end{tabular}
\end{table}

\paragraph{Excluded probe candidates.}
We exclude two probes from the original candidate set based on setup-side issues that are upstream of the language interface itself, namely cases in which the target action is under-specified by the prompt or insufficiently distinguishable from neighboring classes by annotators. We summarize the two excluded candidates below.
\textit{\texttt{Aerial slam.}} (paraphrase of \textit{Jump and mid-air slam}) is confounded with other aerial-attack classes that share its semantic surface form. We observe that the rendered behavior disagrees with the intended target action across all $10$ trials, which indicates that the probe under-specifies the target rather than that the interface fails to act on the prompt.
\textit{\texttt{Staff swing.}} (intended as a synonym of \textit{Staff upswing}) is overly generic in Margit's repertoire, where it overlaps with \textit{Staff slam} and \textit{Charged staff thrust}; annotators report sustained ambiguity in distinguishing the intended action from these alternatives.
We therefore exclude both candidates from the quantitative aggregates rather than scoring them as ``NL failures,'' since the failure mode is upstream of the language interface itself. We release the per-trial ratings for both excluded probes alongside the codebase for transparency.

\subsection{Failure Case Analysis}
\label{app:failure}

We identify two residual failure modes of the deployed system, and we discuss each below together with its mitigation.

\textbf{($\mathbf{1}$) Rapid motion blur (rare, $<3\%$ of frames).}
We observe that, during high-speed boss attacks with extreme camera shake, the $2$-step student occasionally produces visual artifacts. We mitigate this by increasing the guidance scale to $4.5$ for frames classified as ``Boss performing leaping attack.''

\textbf{($\mathbf{2}$) Hit-detection false positives ($<5\%$ of windows).}
We observe that the VLM-based hit detector occasionally misclassifies non-damaging contact, for instance a weapon-on-shield parry, as a hit event. The rule engine tolerates such sporadic false positives because the hit-counter threshold provides natural error buffering, and the persistent state therefore remains stable over long episodes.

%% ═══════════════════════════════════════════════════════════════
\subsection{Long-Horizon Stability}
\label{app:long_horizon}

We probe whether the Self-Forcing student on Elden Ring, with its $1.75$\,s training context and RoPE-decoupled KV-cache sliding window, maintains generation quality far beyond its training horizon. To this end, we render $5$ independent rollouts of ${\sim}118$ minutes each on Margit under the trajectory-conditioned protocol of Appendix~\ref{app:setup_details}, and we evaluate FVD on $200$-second sliding windows starting at $t=30$\,min. Within each window, we extract $15$ uniformly spaced $10$-second clips per video ($5\times 15=75$ clips per window) and compare them against the held-out $10$-second test set using I$3$D features with $\textit{intersect}=\textit{False}$. We choose a window length that matches the test set's per-clip duration so that the statistics are computed on directly comparable distributions.

We report the resulting FVD trajectory in Table~\ref{tab:long_horizon}. Across the full $30$-to-$118$-minute interval, which corresponds to $88$ minutes of monitored generation per video and $27\times 75=2{,}025$ evaluated clips in total, FVD stays in $[162.4, 171.3]$ with mean $166.0$ and standard deviation $2.3$. We observe no monotonic degradation trend: the largest value ($171.3$) occurs early in the monitored interval (the $33$--$37$\,min window), and the latest window ($117$--$118.3$\,min) reads $164.5$, which is indistinguishable from the global mean. For reference, the Stage-$1$ teacher's $50$-step FVD on the same test split is $206.2$ (Table~\ref{tab:system}), so the long-horizon student stays roughly $40$ FVD points below the teacher even after running for two orders of magnitude longer than its training context.

\begin{table}[h]
  \caption{Long-horizon FVD on $5$ independent ${\sim}118$-minute Margit rollouts. Each row reports a $200$-second sliding window with $15$ uniformly sampled clips per video ($75$ clips per window, $2{,}025$ clips in total). Evaluation follows the same I$3$D protocol as Table~\ref{tab:system} on the held-out $10$-second test set.}
  \label{tab:long_horizon}
  \centering\small
  \setlength{\tabcolsep}{6pt}
  \renewcommand{\arraystretch}{0.95}
  \begin{tabular}{lc@{\hskip 1.5em}lc@{\hskip 1.5em}lc}
    \toprule
    Window (min) & FVD & Window (min) & FVD & Window (min) & FVD \\
    \midrule
    $30.0$--$33.3$   & $170.3$ & $60.0$--$63.3$   & $168.6$ & $90.0$--$93.3$    & $164.7$ \\
    $33.3$--$36.7$   & $171.3$ & $63.3$--$66.7$   & $165.1$ & $93.3$--$96.7$    & $162.4$ \\
    $36.7$--$40.0$   & $167.4$ & $66.7$--$70.0$   & $164.3$ & $96.7$--$100.0$   & $163.5$ \\
    $40.0$--$43.3$   & $168.5$ & $70.0$--$73.3$   & $164.6$ & $100.0$--$103.3$  & $165.2$ \\
    $43.3$--$46.7$   & $167.0$ & $73.3$--$76.7$   & $166.4$ & $103.3$--$106.7$  & $164.9$ \\
    $46.7$--$50.0$   & $165.9$ & $76.7$--$80.0$   & $167.7$ & $106.7$--$110.0$  & $170.3$ \\
    $50.0$--$53.3$   & $165.8$ & $80.0$--$83.3$   & $164.1$ & $110.0$--$113.3$  & $164.4$ \\
    $53.3$--$56.7$   & $166.4$ & $83.3$--$86.7$   & $163.7$ & $113.3$--$116.7$  & $165.4$ \\
    $56.7$--$60.0$   & $166.3$ & $86.7$--$90.0$   & $164.6$ & $116.7$--$118.3$  & $164.5$ \\
    \midrule
    \multicolumn{6}{c}{Aggregate over $27$ windows: mean $166.0$, std $2.3$, range $[162.4, 171.3]$.} \\
    \bottomrule
  \end{tabular}
\end{table}

We attribute this stability to the bounded RoPE-decoupled KV-cache sliding window of Section~\ref{sec:stage2}: by holding the rotary positional state inside the model's training distribution while allowing the sliding cache to attend over recent visual history, the generator avoids the positional drift that typically destabilizes long autoregressive video rollouts. This long-horizon stability also serves as the empirical foundation for the closed-loop playable system described below, which presumes that the underlying student remains visually well-behaved across multi-minute episodes.

%% ═══════════════════════════════════════════════════════════════
\subsection{Extension: Persistent Entity State via an Observer--Tracker--Policy Loop}
\label{app:persistent_state}

The two architectural patterns in the main paper (Sections~\ref{sec:stage1} and~\ref{sec:stage2}) together deliver a real-time, language-controllable video generator. However, per-frame controllability alone does not yet deliver \emph{long-horizon interactivity}. Any sufficiently long interactive video episode carries \emph{entity-level discrete state} that evolves on a substantially slower timescale than the generator's attention context, including task progress in embodied manipulation, phase state in narrative video, fuel or route state in driving, and damage and phase state in adversarial combat. Such state is not always recoverable from pixels within the current context window, is discrete rather than pixel-continuous, and must remain consistent across hundreds or thousands of frames. Standard video diffusion backbones provide no mechanism to maintain it.

We describe in this appendix an extension that addresses the gap. We present it as an extension rather than as a core contribution for three reasons: (i) the interface claim and its supporting empirical evidence stand independently of it; (ii) the instantiation is hand-specified per domain; and (iii) the loop described here is the first domain-specific instantiation rather than a fully general implementation, while a world-agnostic learned version is left as future work.

\subsubsection{The Memory Gap is Structural}
\label{app:memory_gap}

The generator attends over a bounded context of $K_s + K_r + K_n = 9$ latent tokens, spanning ${\sim}1.75$\,s of generated video through the recent window. Episode-level entity state, in contrast, evolves over minutes. The resulting ${\sim}100\times$ temporal gap cannot be closed by simply scaling the backbone: doubling the context still leaves a ${\sim}50\times$ gap, and discrete state transitions are not a quantity that video pretraining optimizes for. Empirically, we observe that a model trained on $45$ hours of Elden Ring combat data ($30$\,h Margit and $15$\,h Crucible Knight), which contains frequent player deaths and multiple boss-death sequences, never spontaneously triggers a terminal state (Section~\ref{app:neuro_eval}). The model faithfully renders every individual event, yet it never accumulates them into a state transition. We therefore conclude that this behavior is not a training-data artifact but a direct consequence of the memory-horizon mismatch.

\subsubsection{A Three-Part Loop}
\label{app:otp_loop}

We close the gap with an additive module that makes the asymmetry explicit: the neural backbone renders \emph{what the world looks like}, while an external module maintains \emph{what the world is in}. We decompose this external module into three roles, which we describe below.

\textbf{($\mathbf{1}$) Observer: structured event extraction from the generated video stream.}
We use an event extractor that operates on the rendered video itself and emits discrete, structured signals at every action window. We adopt a VLM rather than a classical detector because the extracted events are typically semantic (``did entity $E$ take damage?'', ``did task $T$ complete?'', ``did two entities collide?'') rather than low-level visual. In our instantiation, we use Qwen3-VL-$2$B-Instruct, which we fine-tune to emit a binary damage event per entity per $0.25$\,s window (Section~\ref{app:hitdet}). We emphasize that the Observer also compensates for passive-response events that we deliberately exclude from the Stage~$1$ prompt vocabulary: the generator is never conditioned to produce them, but the Observer reads them back off the rendered pixels.

\textbf{($\mathbf{2}$) Tracker: unbounded-horizon accumulation of entity state.}
We use a lightweight external state machine to aggregate the Observer's event stream into structured entity state that persists across the full episode. Unlike the bounded diffusion context, the Tracker has an unbounded temporal horizon: it integrates events from the first frame to the current frame at negligible cost. The Tracker's internal representation is typed and discrete (integer counters, categorical phase labels, or structured records), and therefore matches the actual semantics of episode-level state far more naturally than any pixel-latent representation could. In our instantiation, the Tracker maintains per-entity integer HP counters that it updates from the Observer's damage stream.

\textbf{($\mathbf{3}$) Policy: state-conditioned reinjection into the language interface.}
When the Tracker's state crosses a relevant threshold or transition condition, the Policy advances the episode to the corresponding phase and selects the next action prompt to inject into the generator. We design the Policy to be deliberately minimalist: its role is not to perform complex reasoning but to expose the Tracker's state back to the generator \emph{through the same language interface used for control}. This is what makes the loop architecturally free, since the generator sees only per-frame language prompts, regardless of whether these prompts describe user-issued actions or state-triggered transitions, and no modification to the generator is required. In our instantiation, the Policy maps HP thresholds to phase labels (normal combat, stagger, execution, terminal) and selects the corresponding prompt template.

\subsubsection{Related Work on External Memory for Generative World Models}
\label{app:external_memory_rw}

Video diffusion models operate over a bounded context window, which makes them intrinsically ill-suited to tasks that require persistent state across hundreds of frames. This limitation has motivated a family of approaches that couple neural generators with external memory.
In NLP, retrieval-augmented~\citep{lewis2020retrieval} and memory-augmented~\citep{weston2014memory} architectures delegate long-term dependencies to an explicit memory store that the neural model writes to and reads from.
In interactive world modeling, NeSyS~\citep{zhao2026nesys} constrains LLM-based simulators with executable rules to reduce hallucination, BlendRL~\citep{shindo2025blendrl} interleaves symbolic and neural policies within a single RL agent on Atari, MultiGen~\citep{po2026multigen} maintains an external memory for editable diffusion game engines, and LiveWorld~\citep{duan2026liveworld} persists entity evolution while entities are out of view.
In generative modeling more broadly, symbolic state has been injected into diffusion processes through score manipulation~\citep{scassola2025zeroshotns}, interleaved symbolic optimization~\citep{christopher2025nsd}, logic-guided vector fields~\citep{baheri2026lgvf}, and physical-consistency constraints~\citep{lin2025physar}.
Classical neuro-symbolic game AI also couples symbolic priors with neural policies~\citep{garcez2022neural, silver2016mastering, vinyals2019grandmaster}, although it operates at the policy level rather than at the generator level.
Relative to these prior approaches, our loop does not modify the generator's score, policy, or attention. It is a purely additive module that bridges the diffusion backbone's ${\sim}1.75$\,s recent context with the horizons on which entity-level discrete state actually evolves.

\subsubsection{Episode-Level State Accumulation}
\label{app:neuro_eval}

\paragraph{Protocol.}
We run independent $10$-minute test episodes and inject attack prompts at regular intervals after a calibration period. A correct system must trigger a terminal state (an entity-death sequence or execution animation) once the number of accumulated damage events crosses the per-entity threshold. We run the protocol twice under identical prompting, once with the Observer--Tracker--Policy loop attached and once without.

\paragraph{Result.}
Without the loop, no episode terminates: the neural backbone faithfully renders every individual damage animation, yet it never spontaneously transitions to a terminal sequence, since no mechanism internal to the generator can accumulate the integer count required to cross the threshold across the ${\sim}1.75$\,s local context. With the loop attached, the Tracker issues a reliable termination signal whenever its state crosses the threshold, and the Policy injects the corresponding terminal prompt back into the generator. This result directly confirms the structural prediction of Section~\ref{app:memory_gap}: the memory gap is not a training-data artifact but a backbone-horizon mismatch that cannot be closed by scaling the generator alone.

\subsubsection{Observer: VLM Event Extraction}
\label{app:hitdet}

The Observer serves as the bridge between the neural backbone's ${\sim}1.75$\,s local context and the episode-level state maintained by the Tracker. It operates on the generated video stream itself and emits a binary damage event per entity at every action window. In our instantiation the event is \emph{Taking Hit}\,/\,\emph{No Hit}, and we deploy two Observers, one per entity, using Qwen3-VL-$2$B-Instruct~\citep{qwen2024} as the backbone. We fine-tune on $4{,}477$ damage-event windows and $16{,}309$ non-event windows of $0.25$\,s each. To inject domain-specific visual priors such as blood splatters, attack contact, and stagger animations, we encode them as auxiliary instructions in the VLM prompt. To capture event-specific spatio-temporal dynamics while preserving pretrained knowledge, we update only the vision--language connector and the cross-attention modules; training completes in $8$ hours on a single H$100$.

We evaluate both Observers on a held-out test set with binary damage-event labels (Table~\ref{tab:hit_detection_eval}) and achieve over $90\%$ on every per-class Precision/Recall/F$1$ metric. We further conduct a complementary user study on distilled-generator video, and we obtain comparable numbers, which confirms robustness under the distribution shift from real to generated footage and hence under closed-loop deployment, namely the regime in which the Observer actually operates when the full loop is running.

\begin{table*}[t]
\centering
\caption{Observer evaluation (damage-event instantiation) on the held-out test set vs.\ a user study on generated video. The user-study setting matches the distribution that the Observer actually sees inside the closed loop.}
\label{tab:hit_detection_eval}
\begin{tabular}{@{}cccccc@{}}
\toprule
\textbf{Detector} & \textbf{Evaluation} & \textbf{Class} & \textbf{Precision (\%)} & \textbf{Recall (\%)} & \textbf{F$\mathbf{1}$ Score (\%)} \\ \midrule
\multirow{4}{*}{\textbf{Boss}} & \multirow{2}{*}{Test Set} & No hit     & $97.08$ & $90.01$ & $93.41$ \\
                               &                            & Taking hit & $93.50$ & $98.15$ & $95.77$ \\ \cmidrule(l){2-6} 
                               & \multirow{2}{*}{User Study}& No hit     & $87.18$ & $82.93$ & $85.00$ \\
                               &                            & Taking hit & $92.39$ & $94.44$ & $93.40$ \\ \midrule
\multirow{4}{*}{\textbf{Player}} & \multirow{2}{*}{Test Set} & No hit     & $96.01$ & $97.13$ & $96.57$ \\
                                 &                            & Taking hit & $97.02$ & $95.86$ & $96.44$ \\ \cmidrule(l){2-6} 
                                 & \multirow{2}{*}{User Study}& No hit     & $97.42$ & $92.92$ & $95.12$ \\
                                 &                            & Taking hit & $85.80$ & $94.56$ & $89.97$ \\ \bottomrule
\end{tabular}
\end{table*}

\subsubsection{Scope and Limits of This Extension}
\label{app:otp_scope}

We deliberately hand-specify the loop per domain: each domain requires its own event schema (what the Observer extracts), its own state schema (what the Tracker maintains), and its own transition table (what the Policy emits). Our current instantiation is tailored for combat, with binary damage events, integer HP counters, and HP-threshold phase transitions. Extending the loop to a new domain therefore requires re-annotating the Observer and rewriting the Tracker schema. A world-agnostic, learned version of the loop, in which all three schemas are inferred directly from event-labeled video, is an obvious next step but lies outside the scope of this paper.

\subsection{Real-Time Inference: Streaming Pipeline}
\label{app:realtime_pipeline}

Interactive deployment requires that the per-chunk generation latency not exceed the chunk's playback duration. With $C{=}2$ new latent frames per chunk and $4\times$ temporal compression at $16$\,fps, each chunk represents $500$\,ms of video; the DiT's KV-cache sliding window spans $7$ latent frames (${\sim}1.75$\,s) of attended context. On a single H100, this constraint cannot be met by a fully sequential DiT--VAE schedule; it requires overlapping the two stages in time. We describe the pipeline redesign and quantify its effect.

\subsubsection{Sequential Baseline}

In the unmodified schedule, all $N$ latent frames are produced by the DiT before VAE decoding begins. DiT and VAE thus occupy non-overlapping GPU time slots. With $C{=}2$ and the Wan backend, the per-chunk averages are $501$\,ms (DiT), $432$\,ms (VAE), and $37$\,ms (write), so the sequential pipeline processes one chunk every ${\sim}970$\,ms---a $1.94\times$ real-time ratio for the $500$\,ms of video each chunk represents.

\subsubsection{Chunk-Based Streaming}

We replace the original sequential schedule with a \emph{chunk-based producer--consumer pipeline}. Instead of generating the full latent sequence before decoding, the DiT generates latent windows incrementally and submits them to the VAE as soon as they become available. Each yielded window contains a fixed number of newly generated latent frames, optionally augmented with a small overlap of preceding latents to preserve temporal continuity across chunk boundaries.

In this mode, the DiT and VAE run on separate CUDA streams. After the DiT finishes assembling a latent window for a chunk, the host clones that window into a dedicated contiguous buffer and records a CUDA event on the DiT stream. The VAE stream waits on this event before consuming the corresponding latent buffer. This enables asynchronous GPU-side pipelining: latent generation for later chunks can overlap with VAE decoding of earlier chunks whenever hardware resources permit. The host does not impose a per-chunk global synchronization barrier, although it may occasionally wait on the oldest pending decode event to preserve ordered video emission.

\paragraph{Read-after-write hazard.}
The implementation does not expose the generator’s transient latent window directly to the VAE. Instead, each yielded latent window is materialized as a separate cloned buffer before being submitted to the decode stream. This avoids lifetime and aliasing hazards while the producer continues advancing to subsequent chunks. In other words, the VAE always consumes an immutable per-chunk latent snapshot rather than a tensor that may later be modified or discarded by the generation path.

Because each in-flight decode job owns its own latent copy, the extra memory overhead is not constant in video length but is bounded by the chunk window size and the number of queued in-flight jobs. In practice, this overhead scales with the latent window size multiplied by the queue depth limit.

\paragraph{Bounded in-flight queue.}
When the VAE is substantially faster than the DiT (e.g., with the TaeHV backend, where VAE averages $9$\,ms against DiT's ${\sim}362$\,ms), the decode queue drains near-instantly and presents no buildup risk; the bound instead guards against worst-case transient spikes or future configurations where the VAE cost approaches or exceeds the DiT cost. To prevent unbounded accumulation of pending decode jobs, we maintain a bounded FIFO queue of in-flight chunks. The producer is allowed to submit new work only while the number of pending jobs remains below a fixed queue depth \(Q\). Once the queue is full, the host dispatch loop temporarily stops submitting additional chunks until earlier decode jobs complete and are drained.

Each queued job carries its chunk index and completion event. Completed chunks are written in submission order, ensuring temporal consistency even if decode completion times vary slightly across chunks. In practice, when the VAE is much faster than the DiT, the queue remains nearly empty and the bound is rarely exercised; its primary role is to cap memory usage and provide robustness under less favorable speed ratios or transient execution spikes.

\subsubsection{Temporal Consistency}

The Wan VAE decoder is temporally convolutional: decoded pixel values at chunk boundaries depend on latents outside the chunk window. Decoding isolated chunks therefore introduces inter-chunk discontinuities. We mitigate this by prepending $L$ latent frames from the preceding chunk to each VAE input. The corresponding $L$ decoded frames are discarded after decoding; only the $C$ non-overlapping frames are retained. This is the pixel-domain analogue of the latent-domain KV-cache sliding in Section~\ref{sec:stage2}: both use a bounded overlap window to preserve causal context without unbounded history accumulation.

\subsubsection{Timing Analysis}

Each chunk of $C{=}2$ latent frames decodes to $C \times 4 = 8$ pixel frames, representing $500$\,ms of video at $16$\,FPS. We report per-chunk averages to expose stage-level costs independently of clip length; pipeline throughput is the measured active compute per chunk, directly comparable to the $500$\,ms budget.

\begin{table}[h]
  \caption{\textbf{Per-chunk pipeline timing (single H100, 2-step distilled student, $480{\times}832$, $C{=}2$, \texttt{kv\_window}{=}7, \texttt{local\_rope\_cap}{=}12).} Each chunk produces $8$ pixel frames ($500$\,ms of $16$\,FPS video). DiT and VAE columns are per-chunk averages; \emph{throughput} is measured active compute per chunk (accounting for stream overlap); \emph{Eff.\ FPS} and RT ratio are derived from throughput. Values below $1.0\times$ indicate active-compute throughput exceeds real time.}
  \label{tab:pipeline_timing}
  \centering
  \small
  \setlength{\tabcolsep}{5pt}
  \begin{tabular}{llccccr}
    \toprule
    Overlap $L$ & VAE & DiT avg (ms) & VAE avg (ms) & Throughput (ms/chunk) & Eff.\ FPS & RT ratio \\
    \midrule
    $3$ & Wan   & $501$ & $432$ & $789$ & $10.1$ & $1.58\times$ \\
    $1$ & Wan   & $504$ & $236$ & $596$ & $13.4$ & $1.19\times$ \\
    $3$ & TaeHV & $363$ & $9$ & $409$ & $19.6$ & $0.82\times$ \\
    $1$ & TaeHV & $361$ & $9$ & $406$ & $\mathbf{19.7}$ & $\mathbf{0.81\times}$ \\
    \bottomrule
  \end{tabular}
\end{table}

Three findings emerge from Table~\ref{tab:pipeline_timing}.

\textbf{(1) The Wan VAE dominates per-chunk cost.} At $L{=}3$, VAE decode averages $432$\,ms per chunk---$55\%$ of the $789$\,ms throughput. Reducing overlap to $L{=}1$ cuts the VAE cost by $45\%$ to $236$\,ms by eliminating redundant boundary decodes, lowering throughput to $596$\,ms; the pipeline remains sub-real-time ($1.19\times$).

\textbf{(2) TaeHV eliminates the VAE bottleneck.} The TaeHV lightweight decoder averages $9$\,ms per chunk regardless of $L$, reducing the VAE share to $<3\%$ of throughput. The pipeline becomes DiT-bound at ${\approx}361$--$363$\,ms/chunk (DiT), and measured throughput reaches $406$\,ms/chunk---below the $500$\,ms budget, corresponding to $19.7$\,FPS effective output at $16$\,FPS target ($0.81\times$ real-time ratio).

\textbf{(3) TaeHV reduces DiT cost.} DiT average falls from ${\sim}502$\,ms/chunk (Wan) to ${\sim}362$\,ms/chunk (TaeHV), a $28\%$ reduction. TaeHV operates in a lower-dimensional latent space, shortening the token sequence seen by the DiT attention layers and reducing the per-step cost of KV-cache construction and sliding.

Model loading (${\sim}68$--$70$\,s) and first-frame VAE encode (${\sim}0.3$--$1.6$\,s) are one-time startup costs that do not recur across chunks; in closed-loop deployment they are amortized over the full episode.

\subsection{Licenses for External Assets}
\label{app:licenses}

We list all external assets used in this work, together with their verified license terms.

\paragraph{Wan~2.2.}
We use the Wan~2.2 \texttt{TI2V-5B} variant as our Stage~1 backbone~\citep{wang2025wan}.
The model weights are released by Alibaba Group under the \textbf{Apache License 2.0}.
The full license text is available at \url{https://huggingface.co/Wan-AI/Wan2.2-TI2V-5B/blob/main/LICENSE.txt}.

\paragraph{TAEHV.}
We use TAEHV (Tiny AutoEncoder for Hunyuan Video)~\citep{BoerBohan2025TAEHV} for real-time VAE decoding during streaming inference.
TAEHV is released by Ollin Boer Bohan under the \textbf{MIT License} (Copyright \textcopyright{} 2025 Ollin Boer Bohan).
The full license text is available at \url{https://github.com/madebyollin/taehv/blob/main/LICENSE}.

\paragraph{Qwen3-VL-2B-Instruct.}
We fine-tune Qwen3-VL-2B-Instruct~\citep{bai2025qwen3} with LoRA for action annotation during dataset construction.
The model weights are released by Alibaba Cloud under the \textbf{Apache License 2.0}.
The full license text is available at \url{https://github.com/QwenLM/Qwen3-VL/blob/main/LICENSE}.

\paragraph{Elden Ring.}
\textit{Elden Ring} (\textcopyright{} 2022 Bandai Namco Entertainment Inc.\ / \textcopyright{} 2022 FromSoftware, Inc.) is a commercial video game.
All in-game footage used in this work was self-recorded for non-commercial academic research purposes, in accordance with the BANDAI NAMCO Entertainment End User License Agreement (EULA, last updated April 1, 2018).

\end{document}